\documentclass[journal,twoside,web]{ieeecolor}
\usepackage{generic}

\PassOptionsToPackage{
  hidelinks,
  pagebackref=true,
  breaklinks=true,
  colorlinks,
  bookmarks=false
}{hyperref}

\usepackage{float}
\usepackage{amsmath,amssymb,amsfonts}
\usepackage{graphicx}
\usepackage{array}
\usepackage{stfloats}
\usepackage{url}
\usepackage{verbatim}
\usepackage{multirow}
\usepackage{makecell}
\let\labelindent\relax
\usepackage{enumitem}

\usepackage{algorithm}
\usepackage{algorithmic}

\makeatletter
\def\endfigure{\end@float}
\def\endtable{\end@float}
\makeatother

\usepackage[caption=false,font=normalsize,labelfont=sf,textfont=sf]{subfig}

\makeatletter
\let\NAT@parse\undefined
\makeatother
\usepackage{hyperref}
\usepackage{orcidlink}

\usepackage[acronym]{glossaries}
\colorlet{glsblack}{black}

\newcommand{\REV}[1]{\textcolor{black}{#1}}

\def\BibTeX{{\rm B\kern-.05em{\sc i\kern-.025em b}\kern-.08em
    T\kern-.1667em\lower.7ex\hbox{E}\kern-.125emX}}
\usepackage{balance}

\newcommand{\tcm}{TCM}
\newcommand{\AQA}{Action Quality Assessment (AQA)}
\newcommand{\aqa}{AQA}

\newcommand{\CMEAQA}{Cross-view Multimodality Enhanced Action Quality Assessment (CME-AQA)}
\newcommand{\cmeaqa}{CME-AQA}
\newcommand{\AVPF}{Attention-based Visual–Pose Fusion (AVPF)}
\newcommand{\avpf}{AVPF}
\newcommand{\MVA}{Multiscale View Alignment (MVA)}
\newcommand{\mva}{MVA}
\newcommand{\CPR}{Cardiopulmonary Resuscitation (CPR)}
\newcommand{\cpr}{CPR}
\newcommand{\MAE}{Mean Absolute Error (MAE)}
\newcommand{\mae}{MAE}

\providecommand{\refname}{References}

\makeatletter
\def\thebibliography#1{\section*{\refname}%
    \addcontentsline{toc}{section}{\refname}%
    \footnotesize \vskip 0.3\baselineskip plus 0.1\baselineskip minus 0.1\baselineskip%
    \list{\@biblabel{\@arabic\c@enumiv}}%
    {\settowidth\labelwidth{\@biblabel{#1}}%
    \leftmargin\labelwidth
    \advance\leftmargin\labelsep\relax
    \itemsep 0pt plus 1pt\relax%
    \usecounter{enumiv}%
    \let\p@enumiv\@empty
    \renewcommand\theenumiv{\@arabic\c@enumiv}}%
    \let\@IEEElatexbibitem\bibitem%
    \def\bibitem{\@IEEEbibitemprefix\@IEEElatexbibitem}%
    \def\newblock{\hskip .11em plus .33em minus .07em}%
    \if@technote\sloppy\clubpenalty4000\widowpenalty4000\interlinepenalty100%
    \else\sloppy\clubpenalty4000\widowpenalty4000\interlinepenalty500\fi%
    \sfcode`\.=1000\relax}

\makeatother

\begin{document}
\markboth{\hskip25pc IEEE TRANSACTIONS AND JOURNALS TEMPLATE}
{Author \MakeLowercase{\textit{et al.}}: Title}

\title{Cross-view Multimodal Vision-Based Assessment Framework for Traditional Chinese Medicine Rehabilitation Training}

\author{Francis Xiatian Zhang$^{\orcidlink{0000-0003-0228-6359}\ast}$,~\IEEEmembership{Member,~IEEE,}
        Hao Yao$^{\orcidlink{0000-0001-6967-3308}\ast}$,
        Shengxuan Chen$^{\orcidlink{0000-0001-9093-6379}\ast}$,
        Hong Zhu$^{\orcidlink{0000-0003-4964-1610}}$,
        Hongxiao Jia$^{\orcidlink{0000-0002-4924-1388}}$,
        Sisi Zheng$^{\orcidlink{0000-0002-1575-6877}\dag}$, 
        Hubert P. H. Shum$^{\orcidlink{0000-0001-5651-6039}\dag}$,~\IEEEmembership{Senior Member,~IEEE}
\thanks{F. X. Zhang and H. P. H. Shum are with the Department of Computer Science, Durham University, UK (e-mail: \{xiatian.zhang, hubert.shum\}@durham.ac.uk). F. X. Zhang is also with the Institute for Regeneration and Repair, The University of Edinburgh, UK (e-mail: francis.zhang@ed.ac.uk).}
\thanks{H. Yao is with Ningbo Hospital of Traditional Chinese Medicine, Ningbo, China (e-mail: 1242447006@qq.com).}
\thanks{S. Chen is with the Department of Rehabilitation Medicine, The Gulou Hospital of Traditional Chinese Medicine, China (e-mail: 13693012323@163.com).}
\thanks{S. Zheng, H. Zhu, and H. Jia are with \REV{Beijing Key Laboratory of Intelligent Drug Research and Development for Mental Disorders; National Clinical Research Center for Mental Disorders; National Center for Mental Disorders; and Beijing Anding Hospital, Capital Medical University, Beijing, China} (e-mail: \{zhengsisi, zhuhong, jhxlj\}@ccmu.edu.cn).}
\thanks{F. X. Zhang, H. Yao, and S. Chen share co-first authorship ($^{\ast}$) of this paper.}
\thanks{$^{\dag}$Corresponding authors: S. Zheng and H. P. H. Shum.}
\thanks{This is the author-accepted version of an article published in \textit{IEEE Transactions on Neural Systems and Rehabilitation Engineering}. The Version of Record is available at
\url{https://ieeexplore.ieee.org/document/11573058}.}}

\maketitle

\begin{abstract}
Vision-based assessment can provide convenient and cost-effective evaluation in Traditional Chinese Medicine (TCM) rehabilitation training, where action quality assessment (AQA) from computer vision offers a promising solution.
Existing automatic AQA frameworks for physical therapy typically rely on skeletal data captured from a single viewpoint, which is inefficient for TCM techniques such as acupuncture or Tuina that involve dense hand self-occlusion and complex hand–object interactions.
To address these challenges, we propose CME-AQA, a cross-view, multimodal vision-based assessment framework that integrates visual–pose fusion to enhance understanding of environmental context and leverages both first-person and third-person videos during training to improve inference robustness. We collected two dual-view datasets, TCM-AQA61-A (Acupuncture) and TCM-AQA61-T (Tuina), each containing synchronized first- and third-person recordings of 61 subjects with expert annotations.
Experimental results show that our approach \REV{achieves superior or comparable mean performance against competitive baselines}, achieving over 10\% relative improvement in weighted F1 over the best competing method on key rating tasks such as Needle Depth and Quick Needle Insertion, while also reducing mean absolute error in quantitative measures such as insertion time and manipulation frequency. 
Testing on a CPR dataset further demonstrates comparable performance \REV{on several posture-based criteria, suggesting applicability to related structured simulated clinical skill assessments where participant motion is central to evaluation.} 
Overall, CME-AQA enhances assessment accuracy for \REV{structured TCM rehabilitation training} and facilitates more convenient and effective \REV{training-oriented skill evaluation}.
\end{abstract}

\begin{IEEEkeywords}
Action Quality Assessment, Machine Learning, Vision-based Rehabilitation Assessment, Traditional Chinese Medicine
\end{IEEEkeywords}

\begin{figure}[ht]
    \centering
    \includegraphics[scale=0.170]{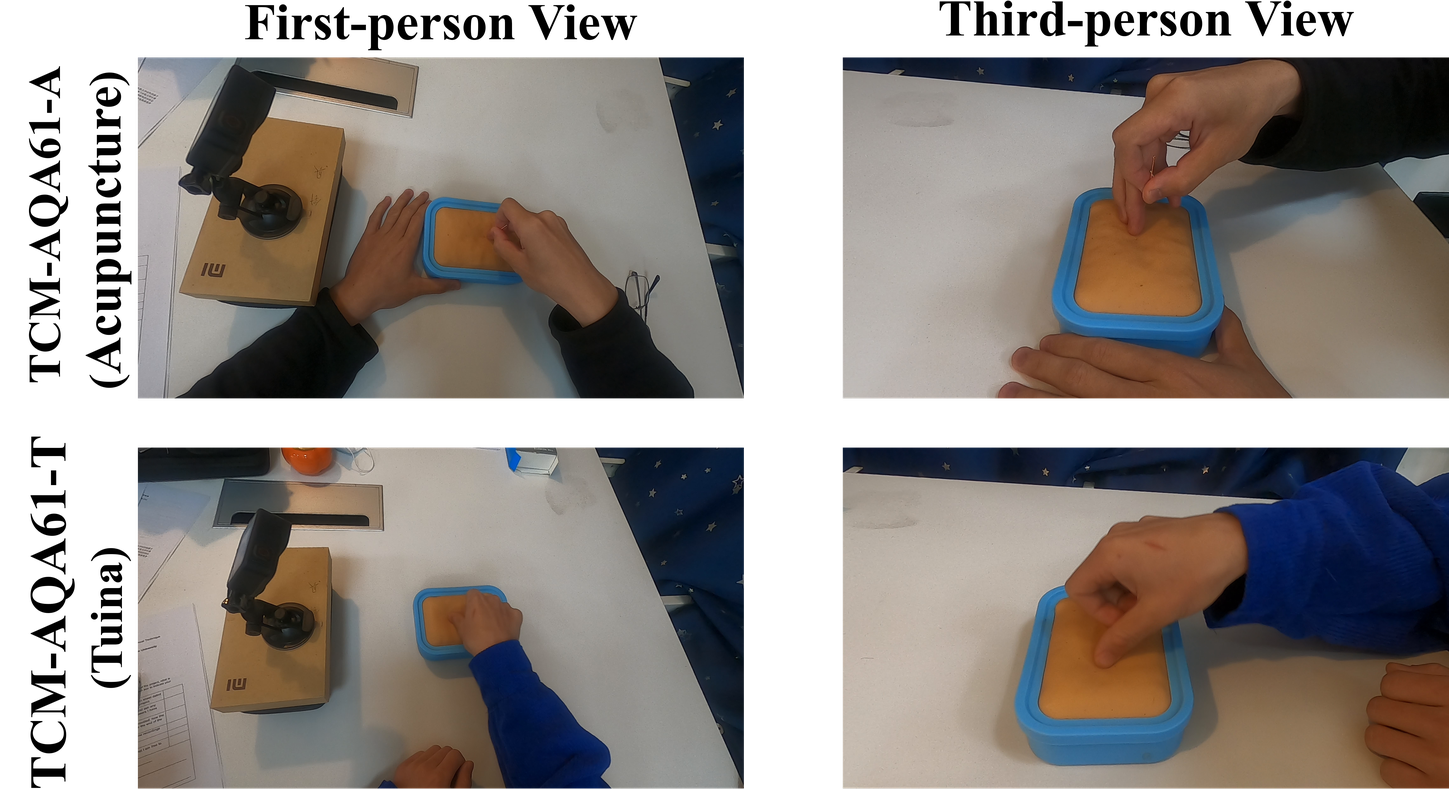}
    \caption{Example frames from our two collected datasets: acupuncture (upper rows) and Tuina (lower rows). Each session was recorded with two cameras: a forehead-mounted first-person view (left column) and a separate third-person view of the hands (right column), providing complementary \REV{participant} and observer perspectives.}
    \label{fig:tcm_data}
\end{figure}

\section{Introduction}
\IEEEPARstart{T}raditional Chinese Medicine (TCM) rehabilitation therapies, such as acupuncture and Tuina (Chinese therapeutic massage), have demonstrated measurable effectiveness in treating motor and neurological disorders \cite{tao2016effects,yu2023effect}.
Recent advancements in TCM rehabilitation training \cite{de2013evidence, du2022mobile, sun2023design} have introduced automatic assessment tools to evaluate \REV{performance in structured TCM rehabilitation training settings}.
These tools enhance accessibility to training, especially in developing regions where experienced instructors are scarce. 
However, many current systems rely on specialized equipment, such as VR glasses \cite{sun2023design} or robotic platforms \cite{du2022mobile}, to assess practice behavior, which limits their practicality. 
To address this limitation, we propose a vision-based assessment framework that derives performance metrics directly from video, providing a cost-effective and portable solution.

In human-performance assessment \cite{rahman2022ai}, action quality assessment (AQA) is a widely adopted method for evaluating practice actions across multiple predefined aspects \cite{liu2024vision}. It applies computer vision techniques to quantify the quality of movements performed during training and has proven effective in both sports and rehabilitation contexts \cite{wu2022survey, rahman2022ai}. Most existing frameworks rely on pose-based skeletal data obtained through sensor-based or markerless motion capture to represent body configurations and dynamics, which helps reduce environmental noise and capture motion details accurately \cite{zhang2023standards}. However, current AQA methods \cite{liao2020deep,deb2022graph,zheng2023skeleton} mainly analyze full-body movements from a single-view perspective, which presents a major limitation for TCM rehabilitation tasks. These procedures require precise hand movements that are frequently obscured by self-occlusion \cite{karvounas2023dynamic}. Moreover, using only skeletal data omits key environmental cues that are critical for modeling hand–object interactions, for example, the manipulation of acupuncture needles on practice pads \cite{jang2022trends}.

To design an effective assessment framework for TCM rehabilitation training, we identify three key challenges in existing AQA methods.
First, current approaches~\cite{liao2020deep,deb2022graph,yao2023contrastive} primarily depend on pose estimation data, which often fail to capture the environmental context where the hands interact with objects and surfaces during TCM procedures. This limitation reduces the capacity of AQA models to evaluate all clinically relevant aspects of rehabilitation performance.
Second, existing AQA frameworks \cite{xu2022finediving,zhang2023logo}, typically designed for single-view video inputs, struggle to handle frequent self-occlusions in which one hand or tool obscures another. Such occlusions can compromise the reliability and continuity of performance evaluation.
Third, to the best of our knowledge, no publicly available dataset currently exists for AQA in TCM rehabilitation \cite{liu2024vision}, making it difficult to establish a standardized benchmark and hindering progress in this research area.

To address these challenges, we propose the \CMEAQA~framework for Traditional Chinese Medicine rehabilitation training. 
The framework integrates both visual and skeletal information from multiple viewpoints to comprehensively capture \REV{participant} movements and their interactions with the environment. 
Attention-based fusion~\cite{zhang2025pose} and multi-view learning~\cite{dong2024lucidaction} have been explored in prior AQA studies. 
CME-AQA advances TCM rehabilitation assessment through a task-specific integration of these mechanisms, together with dedicated dual-view benchmark datasets for this domain.

The \cmeaqa~framework comprises two main components.
The first is the \AVPF~module, which fuses visual and pose features through an attention mechanism to generate a unified and informative representation of motion dynamics and contextual cues. 
The second is the \MVA~training strategy, which leverages both first-person and third-person videos during training to establish multi-view awareness while allowing single-view inference during deployment. 
In addition, to address the lack of publicly available datasets and to establish a benchmark for this task, we created two synchronized multi-view datasets, shown in Fig.~\ref{fig:tcm_data}: TCM-AQA61-A (Acupuncture) and TCM-AQA61-T (Tuina). 
Each dataset includes expert-annotated ratings from 61 subjects, recorded simultaneously from first-person and third-person viewpoints, providing a valuable resource for future research in TCM rehabilitation assessment.

With the proposed datasets, we conducted a comprehensive benchmark study comparing existing action quality assessment methods for physical therapy and validating the effectiveness of our \cmeaqa~framework. 
The experiments demonstrate that our approach achieves over 10\% relative improvement in weighted F1 over the strongest competing baseline on key tasks such as Needle Depth and Quick Needle Insertion, while maintaining competitive performance across other metrics.
These results highlight the capability of the \cmeaqa~framework to enhance TCM rehabilitation training through more precise and reliable measurement. 
Furthermore, our method reduced error by approximately 4\% in metrological evaluations such as insertion time and manipulation frequency, demonstrating its effectiveness in both categorical and continuous assessment tasks. 
To further examine applicability beyond the proposed TCM tasks, we evaluated \cmeaqa~on a \CPR~dataset~\cite{constable2024advancing}, where it achieved measurement performance comparable to human experts \REV{on several posture-based criteria, suggesting applicability to related structured simulated clinical skill assessments where participant motion is central to evaluation.}

The source code and proposed datasets are publicly available on GitHub\footnote{\url{https://github.com/FrancisXZhang/cme-aqa}}. 
For transparency, we note that parts of the Methodology section adapt content from the author's PhD thesis~\cite{zhang2025thesis} and were also presented in the MICCAI~2025 Doctoral Consortium, with permission. 

Our main contributions are summarized as follows:
\begin{enumerate}[leftmargin=*]
\item We propose CME-AQA, a cross-view multimodal action quality assessment framework for \tcm~rehabilitation training, and introduce two synchronized dual-view benchmarks, TCM-AQA61-A (Acupuncture) and TCM-AQA61-T (Tuina), with expert annotations for both categorical and continuous skill indicators, enabling standardized evaluation of fine-grained hand-centric procedures \REV{in structured TCM rehabilitation training settings}.
\item We design an attention-based visual-pose fusion module (AVPF) that performs modality-asymmetric pose-conditioned cross-attention, refining visual representations using pose as a stable conditioning signal to \REV{strengthen interaction-aware RGB-pose fusion for hand-object modeling under self-occlusion in fine-grained TCM procedures}.
\item We introduce a multiscale view alignment (MVA) training strategy that enforces hierarchical cross-view consistency between egocentric and exocentric representations during training, while retaining single-view inference for practical deployment in \REV{routine rehabilitation training settings}.

\end{enumerate}


\section{Related Works}
\subsection{Action Quality Assessment}
\AQA evaluates action execution quality from video by predicting fine-grained performance scores~\cite{wang2021survey}. 
Unlike action recognition, which predicts categorical labels, AQA requires modeling subtle performance variations among visually similar actions. 
Early methods relied on handcrafted motion descriptors or pose-based features~\cite{pirsiavash2014assessing}, followed by convolutional and recurrent architectures for spatiotemporal representation learning~\cite{parmar2019and}.
Recent approaches adopt transformer-based architectures to capture long-range temporal dependencies and structured action dynamics. 
FineParser~\cite{xu2024fineparser} introduces human-centric spatial parsing with step-level temporal decomposition and contrastive regression for score alignment. 
Uni-FineParser~\cite{xu2025human} incorporates mask-supervised attention to enhance human-region modeling while remaining video-only at inference.
Complementary to geometry-aware parsing, PHI~\cite{zhou2025phi} addresses domain shift in long-term AQA through hierarchical feature flow modeling and instruction-guided adaptation, bridging action-recognition pretraining and quality assessment without explicit spatial supervision.

Beyond architectural adaptation, recent work explores richer spatial supervision and multimodal cues for AQA, particularly for geometric reasoning. 
AIFit~\cite{fieraru2021aifit} integrates multi-view RGB with 3D motion-capture and body-shape supervision to provide localized, interpretable feedback. 
PGT~\cite{zhang2025pose} incorporates pose-derived spatial priors and global–local feature decomposition for body-centric modeling. 
LucidAction~\cite{dong2024lucidaction} introduces a hierarchical multi-view and multimodal dataset, where multi-view signals are primarily leveraged for joint training rather than explicit cross-view alignment. 
However, most existing frameworks focus on global full-body analysis, whereas clinical \tcm~procedures require modeling localized hand–object interaction under frequent self-occlusion. 
In such settings, complementary views need to be explicitly aligned to enforce cross-view consistency. 
Without structured alignment, multi-view signals may be underutilized in tasks involving subtle manipulation and strong contextual dependency.

\subsection{Action Quality Assessment in Rehabilitation}
Building upon general AQA frameworks, recent studies have extended assessment to \REV{exercise} training scenarios~\cite{liao2020deep, zahan2024learning, wang2018inertial}.
In the rehabilitation domain, several frameworks~\cite{liao2020deep, yao2023contrastive, zheng2023skeleton} employ AQA-based approaches that analyze human skeletal poses from video to assess movement quality and technique, thereby reducing reliance on specialized kinematic measurement devices~\cite{alamri2008haptic, borghetti2013sensorized}. 
For example, Zheng~\emph{et~al.}~\cite{zheng2023skeleton} compute a dot product matrix across frames to quantify joint rotations in skeletal sequences, enabling more precise modeling of rehabilitation quality. 
More recently, these works have inspired the development of AQA methods for TCM-related skills such as Tai Chi~\cite{li2023taichi, zhao2025ai} and Qigong~\cite{baldinger2025development}, primarily based on human pose representations. 
For instance, Li~\emph{et~al.}~\cite{li2023taichi} reconstruct 3D Tai Chi motion through multi-view pose fusion and quantify motion deviations by comparing \REV{participants’} movements against a reference template. Zhao~\emph{et~al.}~\cite{zhao2025ai} leverage graph neural networks on skeletal representations for Tai Chi movement evaluation.

However, these pose-only frameworks~\cite{liao2020deep, yao2023contrastive, zheng2023skeleton} often overlook visual cues that are critical in complex \tcm~rehabilitation practices such as acupuncture, where key movements involve fine-grained interactions among the hand, needle, and practice pad~\cite{du2022mobile}. This omission limits their ability to capture contextual information necessary for accurate inference. Moreover, reliance on single-view pose estimation makes them vulnerable to self-occlusion, a common challenge in \tcm~therapies characterized by intricate hand motions~\cite{al2017tui}. These limitations highlight the need for a tailored framework that integrates visual–pose fusion and multi-view awareness for reliable assessment of \tcm~rehabilitation performance.

\subsection{Multi-view Clinical Video Analysis}
To mitigate self-occlusion in complex clinical settings, recent video analysis frameworks for clinical applications, such as patient or clinician pose estimation~\cite{xu2022multiview,gerats20233d}, surgical action recognition \cite{schmidt2021multi}, and AQA~\cite{abdelaal2020multi,constable2024advancing}, increasingly employ multi-view camera setups to capture participant movements more comprehensively. These frameworks \cite{xu2022multiview,gerats20233d,schmidt2021multi,abdelaal2020multi,constable2024advancing} typically extract visual or pose features from each video view and fuse them using learnable methods, such as weighted feature fusion across views \cite{constable2024advancing,schmidt2021multi}. For example, Constable et al. \cite{constable2024advancing} proposed a framework that first extracts the pose from each view of \cpr~training videos, then uses graph convolutions to learn joint relationships, and subsequently employs learnable weights to fuse the features of each view to determine the final rating label. However, these methods \cite{xu2022multiview,gerats20233d,schmidt2021multi,abdelaal2020multi,constable2024advancing} depend on complex multi-view setups at inference, which hinders deployment in real-world training environments. Therefore, an effective \aqa~framework for \tcm~rehabilitation should be designed to operate with single-view input during inference while leveraging knowledge learned from multi-view training.

\subsection{Existing Datasets for Rehabilitation AQA}

\begin{table}
\centering
\scriptsize
\caption{Summary of Existing Datasets for Rehabilitation \aqa}
\begin{tabular}{l|c|c|l}
\hline
\textbf{Dataset} & \textbf{Subject} & \textbf{View} & \textbf{Therapy Type} \\ 
\hline
UI-PRMD \cite{vakanski2018data}      & 10   & 1     & Common Rehabilitation Exercise\\ 
KIMORE \cite{capecci2019kimore} &78& 1&Low-back Pain Exercise\\
IntelliRehabDS \cite{miron2021intellirehabds} &29&1& Common Rehabilitation Exercise\\
Keraal \cite{devanne2024medical}   & 31    & 1     & Low-back Pain Exercise  \\ 
FineRehab \cite{li2024finerehab} & 50 & 2 & Musculoskeletal exercises \\ 
\hline
TCM-AQA61-A    & 61    & 2     & Acupuncture  \\ 
TCM-AQA61-T    & 61    & 2     & Tuina   \\ 
\hline
\end{tabular}
\label{tab:datasets}
\end{table}

Most existing rehabilitation \aqa~datasets rely on single-view recordings~\cite{vakanski2018data,capecci2019kimore,miron2021intellirehabds,devanne2024medical}, which limits the ability to train models with multi-view awareness. 
Table~\ref{tab:datasets} summarizes several representative datasets. 
Although the recently introduced FineRehab dataset~\cite{li2024finerehab} includes multi-view recordings, it primarily focuses on full-body patient exercises rather than fine-grained therapist actions. 
Consequently, current datasets are well-suited for general rehabilitation exercises but do not capture the intricate hand movements and frequent self-occlusions typical of \tcm~practices such as acupuncture and Tuina~\cite{jang2022trends}. 
To advance skill assessment in this domain, multi-view datasets specifically designed for \tcm~rehabilitation are essential, enabling models to learn from both \REV{participant} hand motions and contextual interactions with tools and surfaces.

\begin{figure}
    \centering
    \includegraphics[scale=0.11]{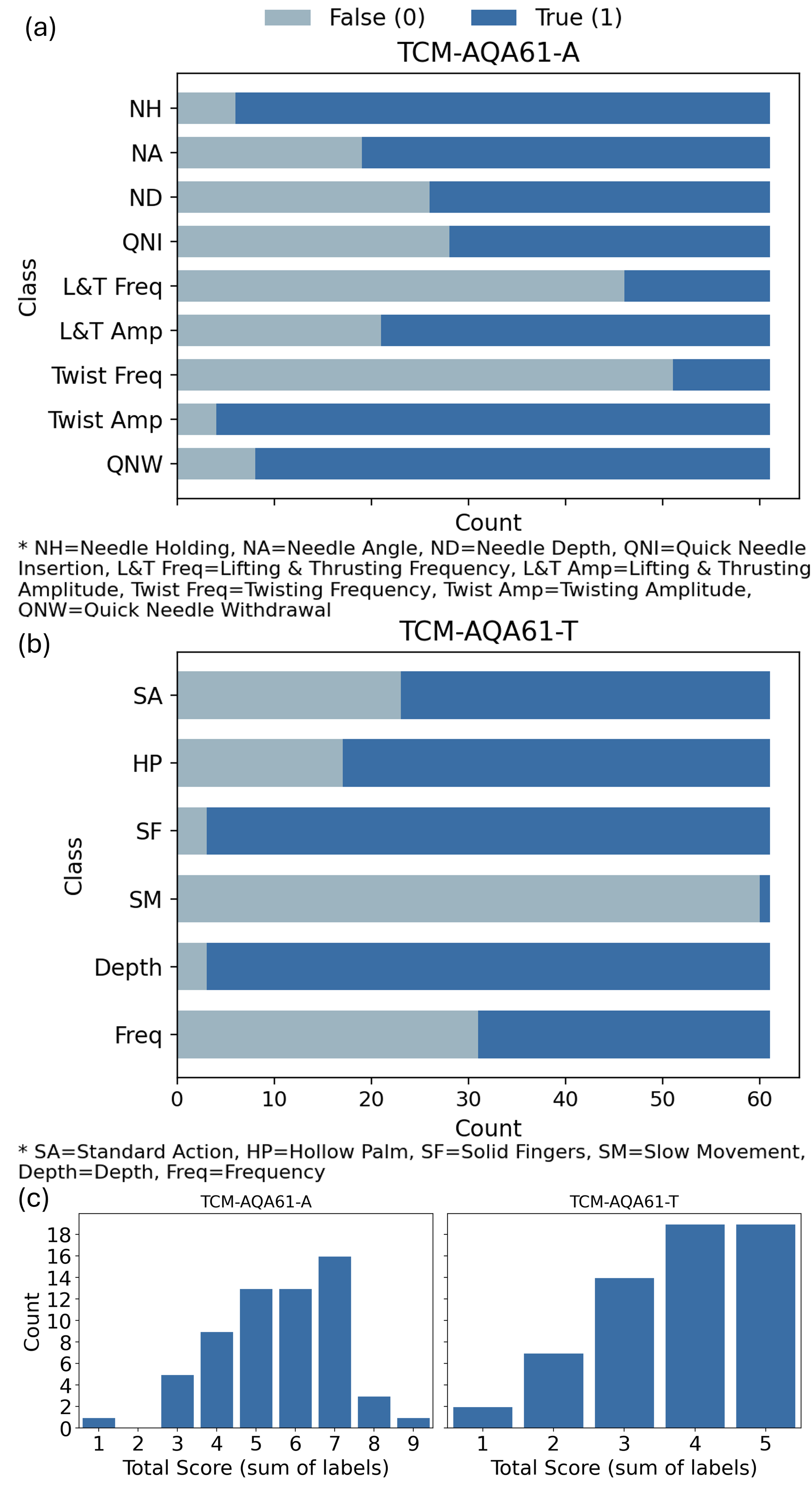}
    \caption{
    Dataset statistics and label distribution for TCM-AQA61-A (Acupuncture) and TCM-AQA61-T (Tuina).
    (a)–(b) Binary distributions of negative (0) and positive (1) annotations for each assessment indicator across 61 subjects.
    Although some indicators exhibit moderate class imbalance (e.g., Twisting Frequency in Acupuncture and Slow Movement in Tuina), no task collapses to a single dominant class.
    (c) Distribution of composite skill scores (sum of binary labels per subject), showing moderate skew toward intermediate-to-high proficiency while preserving variability for multi-aspect evaluation and subgroup analysis.
    }
    \label{fig:hist}
\end{figure}

\begin{figure*}
    \centering
    \includegraphics[scale=0.19]{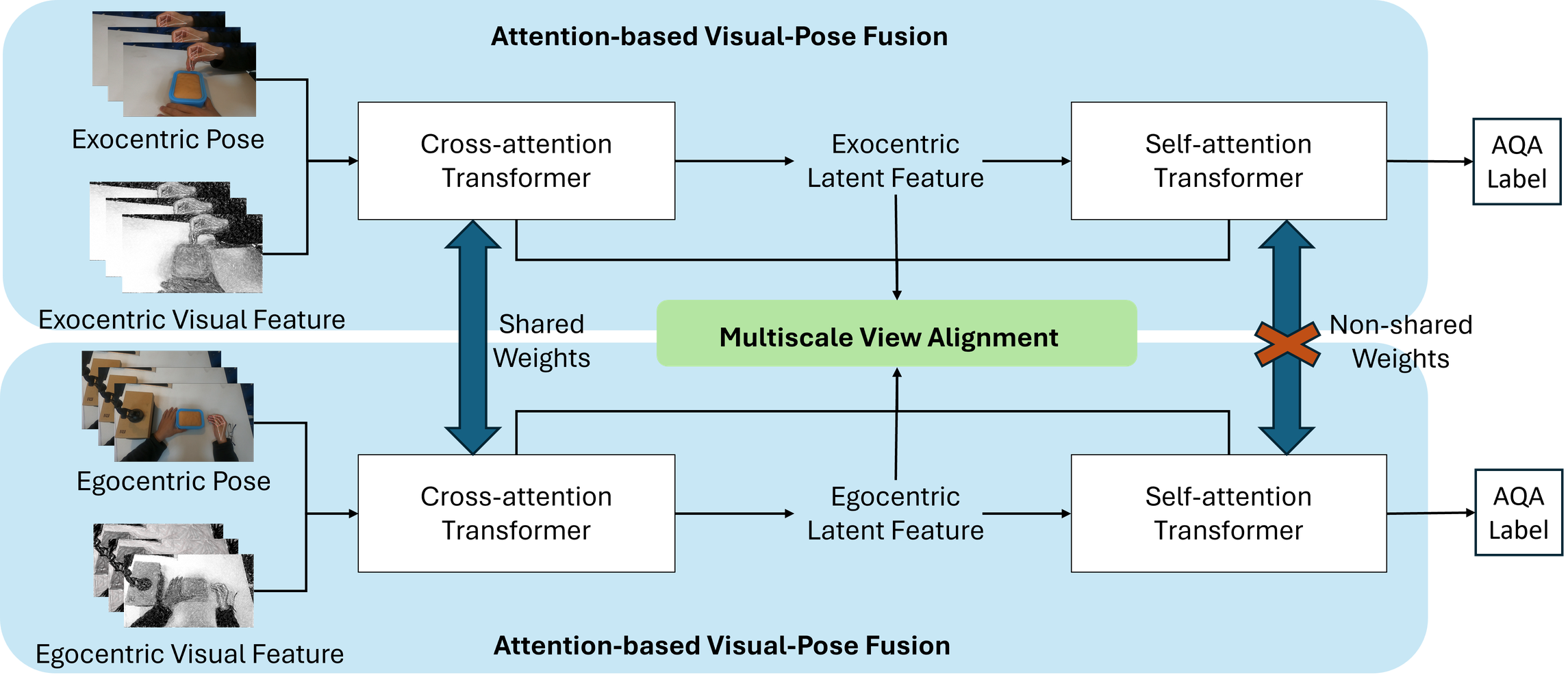}
    \caption{
    Overview of the \CMEAQA~framework, which leverages multi-view and multimodal data for \tcm~rehabilitation assessment. 
    It integrates two main components: (1) the \AVPF~module (Section~\ref{sec:vpf}), which fuses visual and pose features through cross- and self-attention, and (2) the \MVA~training strategy (Section~\ref{sec:mva}), which aligns representations between egocentric and exocentric views to achieve multi-view awareness.
    During inference, only the exocentric (third-person) view is required.
    }
    \label{fig:overview}
\end{figure*}
\section{Data Collection}
\subsection{Procedure}
To the best of our knowledge, no publicly available multi-view video dataset exists for clinical \tcm~rehabilitation training. To address this gap, we introduce two intervention-specific datasets: TCM-AQA61-A (Acupuncture) and TCM-AQA61-T (Tuina). Each dataset contains recordings from 61 subjects, supporting robust model training and evaluation. Both datasets emphasize hand motions, hand–object interactions, and fine motor skills central to \tcm~rehabilitation, providing a foundation for vision-based skill assessment in traditional medicine.

The datasets provide synchronized dual-view recordings of hand movements (Fig.~\ref{fig:tcm_data}), including an egocentric (first-person) and an exocentric (third-person) perspective~\cite{ardeshir2018exocentric}. This configuration captures complementary motion cues from the \REV{participant}’s and observer’s viewpoints. 
The egocentric view emphasizes in-plane hand motion and subtle manipulation dynamics, while the exocentric view better captures vertical displacement and depth progression. 
Compared to conventional fixed multi-camera configurations~\cite{li2024finerehab}, this design preserves fine-grained hand–object interaction together with global spatial context. 
Such complementary perspectives facilitate cross-view representation learning in \cmeaqa.

The data collection process was approved by the Ethical Committee of the Department of Computer Science at Durham University (Reference ID: COMP-2023-03-24T13\_52\_41-slxb76). 
It encompassed acupuncture and Tuina practice sessions on a simulated practice pad, conducted by 61 medical students from the Beijing University of Chinese Medicine, \REV{representing a structured teaching cohort for TCM rehabilitation skill assessment}.
Videos were recorded using two GoPro HERO8 cameras: one mounted on the subject’s forehead to provide an egocentric perspective and another positioned to capture an exocentric view of the hands.

Assessment criteria were defined according to the Clinical Practice Guidelines of Traditional Chinese Medicine~\cite{CACM2021}. 
For TCM-AQA61-A, evaluated aspects included Needle Holding, Needle Angle, Needle Depth, Quick Needle Insertion, Lifting and Thrusting Frequency, Lifting and Thrusting Amplitude, Twisting Frequency, Twisting Amplitude, and Quick Needle Withdrawal. 
For TCM-AQA61-T, evaluated aspects included Sinking Shoulders, Dropping Elbows, Suspended Wrists, Hollow Palms, Solid Fingers, Elbow/Forearm Force, Depth, and Frequency. After data collection, two experienced \tcm~physical therapists independently assessed all subjects using synchronized dual-view recordings, followed by a consensus discussion to resolve discrepancies. 
Inter-rater reliability was computed on the first-round annotations prior to consensus using Cohen’s $\kappa$ and the two-way random-effects intraclass correlation coefficient (ICC(2,1)) across all indicators. The mean $\kappa$ was 0.62, and the average ICC was 0.63, indicating moderate agreement~\cite{landis1977measurement}. These results demonstrate acceptable annotation consistency between experts and support the reliability of the dataset for subsequent model training and evaluation.

In addition to these classification-based assessments, four key aspects were manually annotated with continuous values to support metrological assessment: insertion time (from initial needle–surface contact until stabilization), withdrawal time (from initiation of withdrawal until the needle is fully removed), and manipulation frequency (per second) for both acupuncture and Tuina. These continuous annotations were designed to provide more detailed feedback and enhance the training utility of the system.

\subsection{Dataset Statistics and Label Distribution}
The label distributions of all binary assessment indicators are shown in Fig.~\ref{fig:hist}.
Each sub-skill includes 61 samples with varying proportions of negative (0) and positive (1) labels.
For TCM-AQA61-A, positive samples range from 10 (Twisting Frequency) to 57 (Twisting Amplitude).
Several aspects are relatively balanced (e.g., Needle Depth: 26 vs.\ 35; Quick Needle Insertion: 28 vs.\ 33),
whereas others are moderately skewed due to structured training (e.g., Twisting Frequency: 51 vs.\ 10).
For TCM-AQA61-T, positive counts range from 1 (Slow Movement) to 58 (Solid Fingers and Depth).
Frequency remains nearly balanced (31 vs.\ 30), while Slow Movement is near-saturated (60 vs.\ 1),
reflecting the consistency of this criterion among participants.

Fig.~\ref{fig:hist}(c) further illustrates the distribution of composite skill scores (sum of binary labels per subject). For TCM-AQA61-A, total scores span 1 to 9, with most concentrated between 4 and 7, indicating moderate to high but non-saturated proficiency. For TCM-AQA61-T, scores range from 1 to 5, with most subjects scoring between 3 and 5. Although the distributions are moderately skewed toward intermediate to high performance, consistent with the structured training background of medical students, they retain meaningful variability across skill levels.

Overall, although some sub-skills show moderate imbalance, the dataset is not dominated by a single label, and composite scores retain sufficient spread for multi-aspect evaluation and subgroup analysis. To reduce potential bias, we report both accuracy and class-sensitive metrics such as weighted F1 score in subsequent experiments.

\section{Methodology}
Fig.~\ref{fig:overview} illustrates CME-AQA. At inference, the model takes a single exocentric (third-person) video~\cite{liao2020deep,zheng2023skeleton} and predicts aspect-wise scores.
We extract visual features $F_V\in\mathbb{R}^{T\times C}$ and hand pose features $F_P\in\mathbb{R}^{T\times C}$ per frame, where $T$ denotes the number of frames and $C$ the feature dimension, and fuse them with \AVPF~via pose-conditioned cross-attention followed by temporal self-attention.
The cross-attention module refines visual representations using geometric cues from hand pose, \REV{supporting occlusion-aware refinement when visual cues are available}, while temporal self-attention models long-range dependencies necessary for capturing fine-grained procedural dynamics.
The resulting representation is then mapped to task outputs by a prediction head.

During training, \MVA~uses synchronized FPV and TPV pairs to align multi-scale latent features across views. Multi-scale alignment encourages view-invariant representations at different abstraction levels, enabling robust single-view inference at deployment.

\subsection{\AVPF Module}
\label{sec:vpf}
Conventional \aqa~methods for rehabilitation primarily rely on skeletal representations~\cite{liao2020deep, yao2023contrastive, zheng2023skeleton}, which are insufficient for modeling fine-grained hand–object interaction in \tcm~procedures such as acupuncture~\cite{du2022mobile}. 
Recent multimodal AQA approaches~\cite{zhang2025pose} integrate pose and visual features through pose-guided attention and global–local body modeling, where pose cues primarily bias spatial attention during visual feature extraction. Because pose influences representations only through attention modulation, its geometric information may not be strongly preserved across layers. 
For hand-centric procedures, where pose defines interaction geometry, a more explicit pose–vision coupling is therefore required. To this end, \avpf~performs pose-conditioned refinement by using pose features as keys and values to update visual representations while keeping pose features fixed as a geometric reference.

\begin{figure}
    \centering
    \includegraphics[scale=0.44]{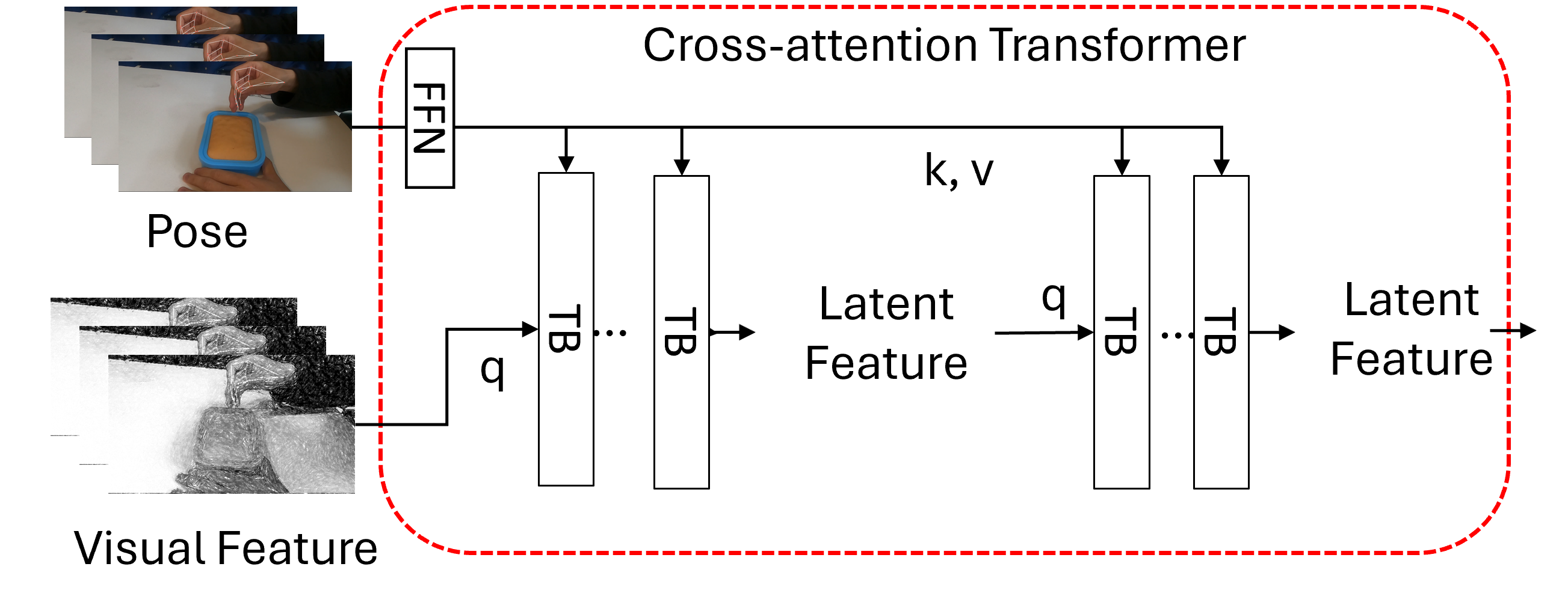}
   \caption{Architecture of the cross-attention transformer in our \avpf~module. Visual features ($Q$) are refined with pose features ($K,V$), while pose features remain fixed. This design enables the model to attend to pose cues relevant to the visual context, \REV{supporting interaction-aware feature fusion for AQA}.}
    \label{fig:ca}
\end{figure}

\begin{figure}
    \centering
    \includegraphics[scale=0.44]{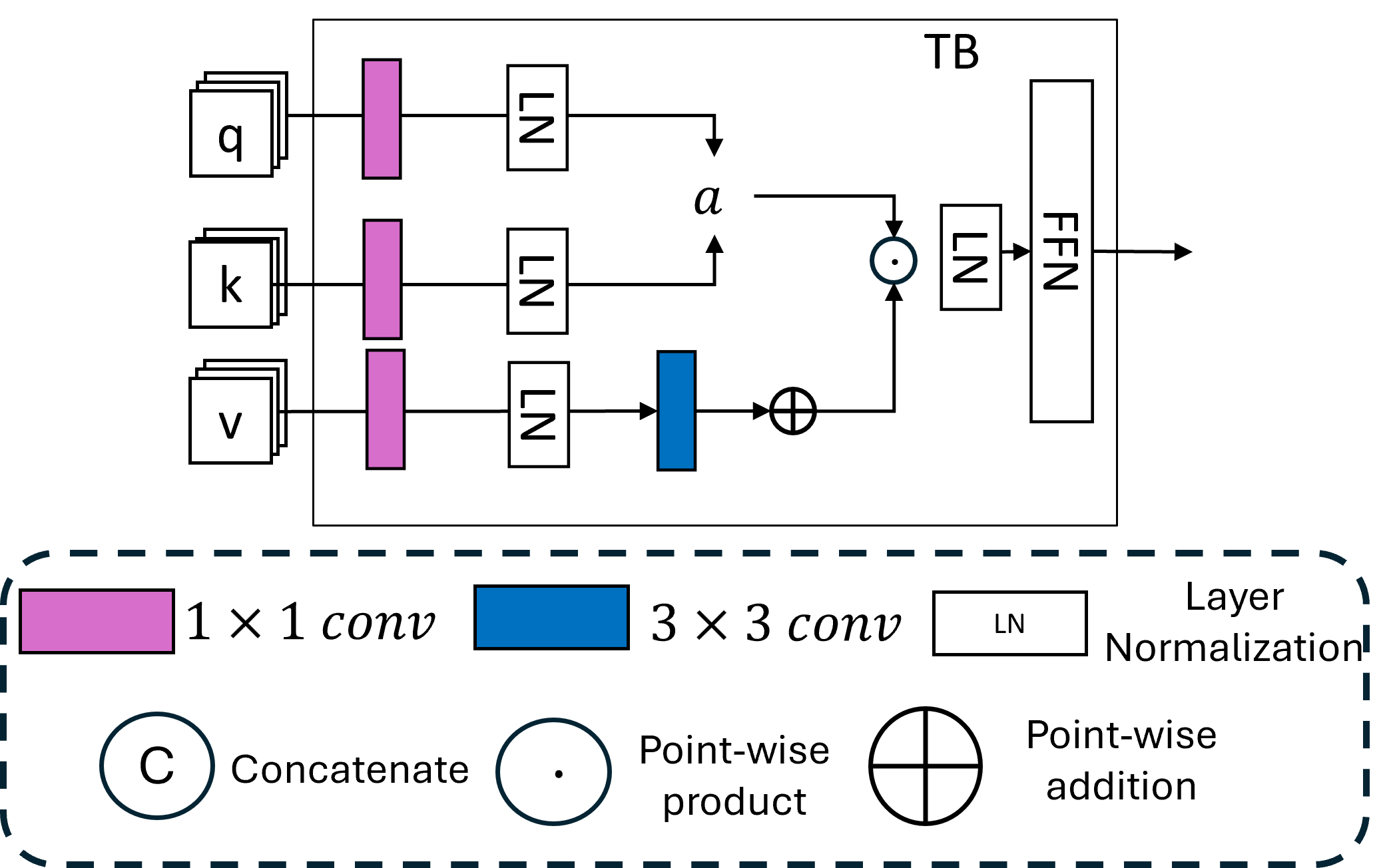}
    \caption{The architecture of our layer-normalization-enhanced attention transformer block. To mitigate potential noise caused by pose estimation inaccuracies, an additional normalization layer is applied after the 1D convolution to enhance robustness.}
    \label{fig:attn}
\end{figure}

\begin{figure}
    \centering
    \includegraphics[scale=0.44]{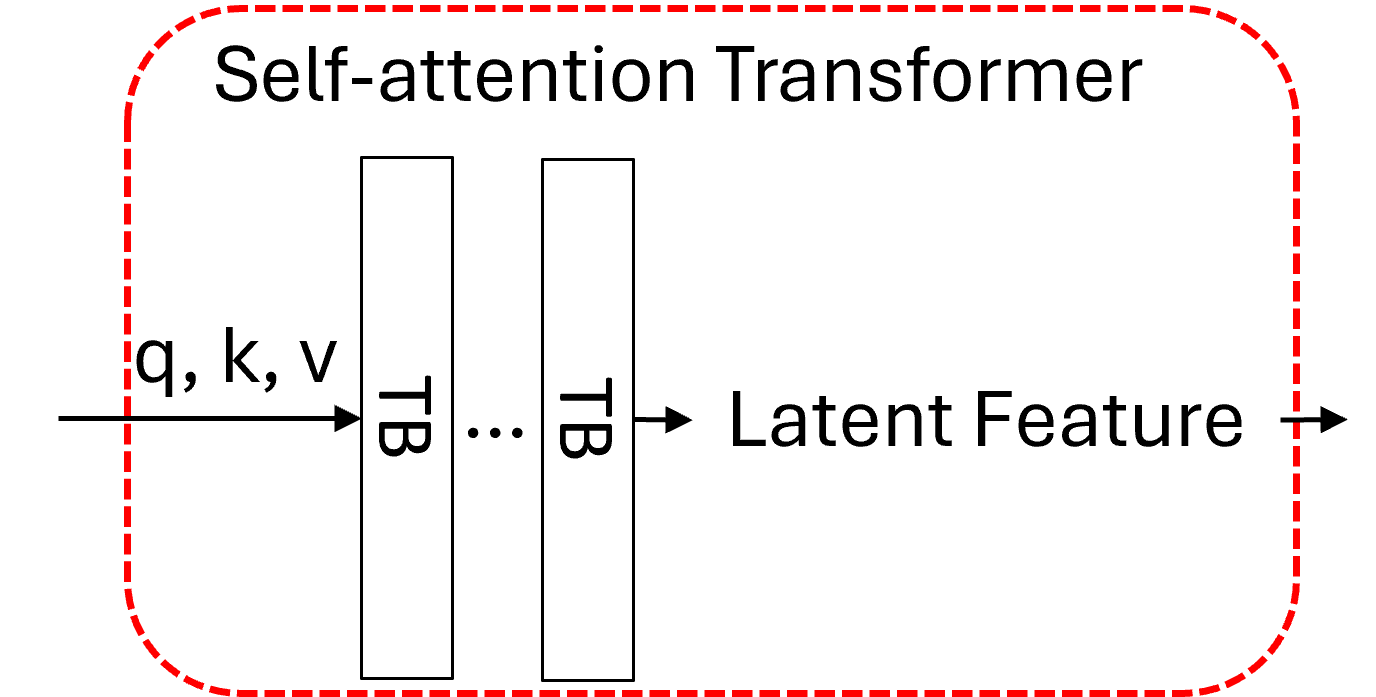}
    \caption{Architecture of the self-attention transformer in our \avpf~module. 
    Fused features act as $Q$, $K$, and $V$, allowing the model to refine representations by focusing on the most relevant information for \aqa~inference.}
    \label{fig:sa}
\end{figure}

As shown in Fig.~\ref{fig:ca}, \avpf~employs a multi-layer cross-attention transformer~\cite{chen2021crossvit} to condition visual features on hand pose. 
Cross-attention enables structured feature-level interaction beyond simple concatenation~\cite{gadzicki2020early}. 
In our formulation, pose features act as keys and values, while only visual features are updated at each layer. This asymmetric update allows iterative refinement of visual representations while preserving stable pose structure.

For implementation, visual features are extracted using a pretrained backbone (e.g., ResNet~\cite{he2016deep}) to obtain $F_V \in \mathbb{R}^{T \times C}$. 
Hand pose features are obtained using a single-view 3D pose estimator (MediaPipe Hands~\cite{lugaresi2019mediapipe,zhang2020mediapipe}) and projected to the same dimensionality, producing $F_P \in \mathbb{R}^{T \times C}$.
Each cross-attention block, indexed by $i$, updates $F_V^i$ using pose features $F_P$ through a layer-normalized attention formulation:
\begin{equation} 
\text{Attn}^i(Q, K, V) = \text{softmax}\left(\frac{LN(Q) LN(K)^T}{\sqrt{d_k}}\right) LN(V),
\end{equation}
where $Q$ is derived from $F_V^i$, and $K,V$ are derived from $F_P$ via 1D convolutions. 
As illustrated in Fig.~\ref{fig:attn}, an additional layer normalization is applied after the 1D convolution to enhance robustness against pose estimation noise. The attended features are then passed through an FFN to produce the updated $F_V^{i+1}$. Stacking $I$ layers yields the fused representation $F_V^I$, while pose features remain unchanged throughout.

The fused representation $F_V^I$ is further processed by a stack of layer-normalized temporal self-attention blocks (Fig.~\ref{fig:sa}) to model intra-sequence dependencies:
\begin{equation}
\text{Attn}^j(Q', K', V') = \text{softmax}\left(\frac{LN(Q') LN(K')^T}{\sqrt{d_{k'}}}\right) LN(V'),
\end{equation}
where $Q', K', V'$ are derived from $F_V^j$ via 1D convolutions. 
Stacking $J$ layers produces the refined representation $F_V^J$.

Finally, $F_V^J$ is passed through a two-layer fully connected network to predict aspect-wise action quality scores.

\begin{figure}[ht]
    \centering
    \includegraphics[scale=0.25]{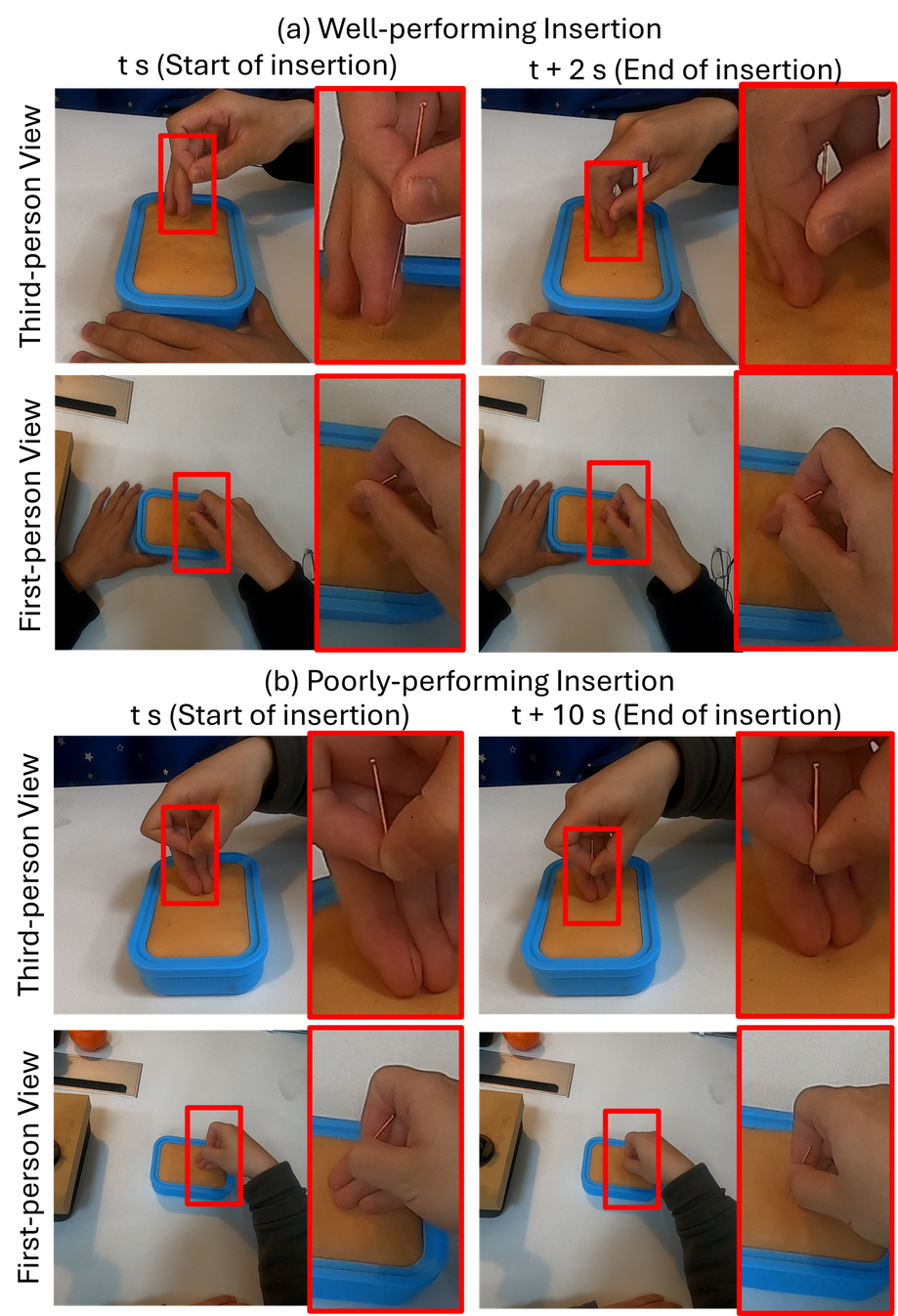}
    \caption{Illustration of complementary motion cues captured by third-person (TPV) and first-person (FPV) views in TCM-AQA61-A during needle insertion. (a) Well-performed trial where the needle is inserted stably and quickly; (b) poorly performed trial where the needle is inserted irregularly and slowly. In each case, the upper row shows TPV frames and the lower row shows FPV frames. TPV better exposes the vertical insertion trajectory relative to the practice pad, while FPV highlights in-plane adjustments of the needle. This cross-view complementarity motivates the proposed \mva~strategy for aligning representations across views during training.}
    \label{fig:fpv_tpv_comparison}
\end{figure}

\subsection{\MVA Training Strategy}
\label{sec:mva}
Conventional \aqa~methods for \tcm~rehabilitation rely on a single exocentric view during both training and inference~\cite{liao2020deep, yao2023contrastive, zheng2023skeleton}. 
However, hand-centric procedures often exhibit view-dependent motion cues~\cite{han2022umetrack}. As shown in Fig.~\ref{fig:fpv_tpv_comparison}, TPV captures vertical insertion and depth progression, whereas FPV highlights in-plane adjustments and subtle needle rotations. Such complementary yet view-specific information cannot be fully represented by a single viewpoint.
While multi-view AQA can mitigate occlusion, existing approaches typically treat synchronized views as independent samples or fuse them via feature aggregation~\cite{dong2024lucidaction,constable2024advancing}. These strategies emphasize information accumulation rather than enforcing cross-view consistency and are not inherently designed for single-view deployment.

To address this limitation, \mva~formulates multi-view learning as a cross-view representation alignment problem. During training, synchronized FPV–TPV pairs are processed through shared cross-attention modules, and latent representations are aligned at multiple hierarchical stages within \avpf.

Specifically, we define three alignment scales: (i) early cross-attention features, where visual and pose representations are initially fused; (ii) \avpf~output features that capture integrated multimodal embeddings; and (iii) self-attention features that provide view-refined representations. At each scale, the corresponding features are projected through a two-layer fully connected network with a nonlinear activation function prior to alignment. The alignment loss is defined as:
\begin{equation}
    L_{\text{Align}} = \sum_{m=1}^{n} \lambda_m \left\| F_{V,\text{exo}}^m - F_{V,\text{ego}}^m \right\|_1,
\end{equation}
where $F_{V,\text{exo}}^m$ and $F_{V,\text{ego}}^m$ denote exocentric and egocentric latent features at the $m$-th scale, and $\lambda_m$ balances contributions across scales. 
This regularization enforces cross-view consistency while preserving hierarchical feature structure.

During training, cross-attention weights are shared across views to capture view-invariant geometric correlations between visual and pose features, whereas self-attention weights remain view-specific to preserve task-adaptive refinement. 
The overall objective is defined as
\begin{equation}
L = \alpha L_{\text{Align}} + \beta L_{\text{AQA}},
\end{equation}
where $L_{\text{AQA}}$ denotes the classification or regression loss for quality prediction, and $\alpha,\beta$ balance alignment and task supervision. 
This formulation enables view-invariant representation learning while retaining discriminative capacity for AQA.

\section{Experiment}
\subsection{Experimental Design}
We conducted experiments on our collected datasets, TCM-AQA61-A and TCM-AQA61-T, as well as on an external multi-view dataset of 40 \cpr~training practices proposed by Constable et al. \cite{constable2024advancing}. The TCM-AQA61 datasets were used for cross-validation and model optimization, while the CPR dataset was used to assess generalization across different practices and view configurations.

\subsubsection{Experiments on TCM-AQA61-A and TCM-AQA61-T}
Our main experiments were conducted on the proposed TCM-AQA61-A and TCM-AQA61-T datasets. Each dataset was split into five folds for cross-validation. Consistent with recent automatic rehabilitation assessment studies~\cite{liao2020deep}, we report results as mean $\pm$ standard deviation across folds to reflect performance variability for each assessment indicator. ResNet \cite{he2016deep} was employed as the backbone for visual feature extraction, and MediaPipe Hands \cite{lugaresi2019mediapipe,zhang2020mediapipe} was used for markerless skeletal extraction, as both are widely adopted in AQA and provide standard visual/skeleton features \cite{zhou2024comprehensive}.

The network was configured with the following hyperparameters: $J = 2$, $I = 2$, $\alpha = 0.5$, and $\beta = 0.5$. We set a batch size of 2 and trained the model for 50 epochs using the Adam optimizer with a learning rate of $1 \times 10^{-4}$. All experiments were conducted on an NVIDIA GeForce RTX 4090 GPU.

For metrological assessments such as insertion time, withdrawal time, and manipulation frequency, we used \MAE to quantify the deviation between predicted values and expert annotations. 
\MAE~is a standard metric for regression tasks and offers a straightforward measure of prediction accuracy in continuous-valued \aqa~\cite{constable2024advancing}. 
Together with classification metrics, it enables a comprehensive evaluation of both categorical and continuous outcomes, supporting precise and actionable feedback in \tcm~rehabilitation training.

For classification assessments, each aspect was evaluated using accuracy and weighted F1 score. 
Accuracy reflects overall prediction correctness, while the weighted F1 score balances precision and recall under class imbalance, providing a more informative metric for clinical data~\cite{chen2024towards}. 
These evaluation criteria are well-suited to clinical applications of \aqa, where reliability and interpretability are essential for effective \tcm~rehabilitation assessment~\cite{wang2021survey}.

\subsubsection{Experiments on CPR Dataset}
To validate the generalization of our method across different practices and view configurations, we evaluated our framework on a multi-view dataset of 40 \cpr~training videos proposed by Constable et al. \cite{constable2024advancing}. The evaluation of our \cmeaqa~framework followed the evaluation protocol of Constable et al. \cite{constable2024advancing}. This dataset includes individual evaluations from two experts as well as the consensus evaluation of the \cpr~practice, covering aspects such as Hand Position, Arm Position, Shoulder Position, Depth, Rate, and Compression Release.

\begin{figure}[h]
\centering
\includegraphics[scale=0.27]{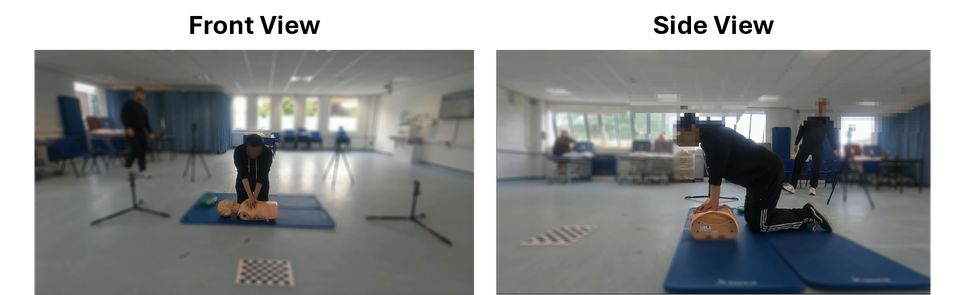}
\caption{Example frames of the views used. For fair comparison, both the front and side views were included during training, as these were the main viewpoints used by experts, while only the front view was used for inference.}
\label{fig:cpr_view}
\end{figure}

Our \cmeaqa~framework was retrained on this dataset. Since the ratings for each subject range from 0 to 4 in this dataset, the \aqa~task in this context is treated as a regression task. To align with this, we modified the final layer of our framework to output a regression prediction. Unlike our collected datasets, which include first-person and third-person views, the CPR dataset contains only front-view and side-view configurations, as egocentric views were not available. Because large portions were blurred for privacy, reliable visual features were unavailable; we therefore used skeleton data only. As a result, only skeletal data could be used as input in our \cmeaqa~framework. To facilitate a fair comparison with expert evaluation, data from both front and side views were utilized during training (as shown in Fig. \ref{fig:cpr_view}), as they are the most commonly chosen views for evaluation in this dataset, while only the front view was used during inference.

The network hyperparameters were configured to match those used in our experiments on our collected datasets to verify the generalization ability. We set a batch size of 2 and trained the model for 50 epochs using the Adam optimizer with a learning rate of $5 \times 10^{-5}$. All experiments were conducted on an NVIDIA GeForce RTX 4090 GPU.

For training and testing, we used a fivefold cross-validation protocol, allocating 80\% of the data for training and 20\% for testing. Following Constable et al. \cite{constable2024advancing}, the \MAE~metric was used to quantify the average error between the predicted scores and the expert-agreed scores.

\subsection{Metrological Assessment Performance}
\begin{table}[ht]
\scriptsize
\centering
\caption{Metrological assessment performance comparison on the TCM-AQA61 dataset. Results are reported as MAE (mean$\pm$SD over folds). For each metric, the method with the lowest mean is highlighted in bold; when means are identical, the method with the lower SD is selected.}
\label{tab:quantitative_performance}
\begin{tabular}{lccc|c}
\hline
\textbf{Method} & \multicolumn{3}{c|}{\textbf{Acupuncture}} & \textbf{Tuina} \\
 & \makecell{Insertion\\ Time (s) ($\downarrow$)}
 & \makecell{Withdrawal\\ Time (s) ($\downarrow$)}
 & \makecell{Frequency\\ (Hz) ($\downarrow$)}
 & \makecell{Frequency\\ (Hz) ($\downarrow$)} \\
\hline
STGCN \cite{yan2018spatial}       & 3.51$\pm$0.91  & 1.05$\pm$0.22 & 0.56$\pm$0.14 & 0.75$\pm$0.14 \\
STNN \cite{liao2020deep}          & 3.26$\pm$1.06  & 0.94$\pm$0.15 &0.53$\pm$0.05 & 0.65$\pm$0.08\\
STGCN-LSTM \cite{deb2022graph}    & 3.55$\pm$0.79 & 0.94$\pm$0.17 & 0.51$\pm$0.05 & 0.65$\pm$0.05 \\
STGCN-RI \cite{zheng2023skeleton} & 6.98$\pm$4.25 & 1.71$\pm$0.78 & 0.93$\pm$0.59 & 0.68$\pm$0.06 \\
FineParser \cite{xu2024fineparser}& 3.44$\pm$1.61 & 1.08$\pm$0.12 & 0.61$\pm$0.22 & 0.69$\pm$0.12 \\
Uni-FineParser \cite{xu2025human}& 3.43$\pm$0.71 &\textbf{0.94}$\pm$0.12 & 0.50$\pm$0.05 & 0.64$\pm$0.07  \\
PGT \cite{zhang2025pose}& 3.45$\pm$0.63 &  0.96$\pm$0.11 & 0.50$\pm$0.06 & 0.66$\pm$0.08 \\
PHI \cite{zhou2025phi}& 3.13$\pm$1.15 &  0.99$\pm$0.21 & 0.52$\pm$0.06 & \textbf{0.63}$\pm$0.07 \\
\textbf{Ours}                     & \textbf{3.12}$\pm$1.04  & 0.94$\pm$0.14 & \textbf{0.49}$\pm$0.05 & 0.64$\pm$0.05 \\
\hline
\end{tabular}
\end{table}

\noindent\textbf{Baselines.}
We conducted a benchmarking study on our dataset to compare the proposed framework with representative baselines. Baselines 1–4 correspond to classical pose-based methods commonly used in rehabilitation-oriented action quality assessment. Specifically, (1) STGCN~\cite{yan2018spatial} is a spatial–temporal graph convolutional network for skeleton-based action modeling; (2) STNN~\cite{liao2020deep} is an early LSTM-based framework that leverages pose sequences for rehabilitation exercise assessment; (3) STGCN-LSTM~\cite{deb2022graph} extends STGCN with an LSTM module to capture longer temporal dependencies; and (4) STGCN-RI~\cite{zheng2023skeleton} incorporates joint rotation matrices into an STGCN backbone to enhance motion representation for rehabilitation skill evaluation. All four baselines were retrained on our dataset under a unified evaluation protocol.

Baselines 5–8 represent more recent transformer-based methods originally developed for general action quality assessment. These include (5) FineParser~\cite{xu2024fineparser}, which uses transformer-based spatio-temporal parsing to align human-centric action steps between query and exemplar videos; (6) Uni-FineParser~\cite{xu2025human}, a unified variant that introduces mask-guided spatial attention and a temporal transformer with learnable queries for fine-grained contrastive score regression; (7) PGT~\cite{zhang2025pose}, a pose-guided transformer framework that integrates pose heatmaps into spatial attention and models global–local body dynamics for fine-grained AQA and (8) PHI~\cite{zhou2025phi}, which applies long-term transformer modeling for absolute score regression from single-view RGB videos with progressive domain adaptation. For FineParser and Uni-FineParser, we approximate the foreground using hand poses and replace the original classification head with a two-layer fully connected regression head to enable metrological prediction, as the original implementations are classification-oriented.

\noindent\textbf{Results.}
As shown in Table~\ref{tab:quantitative_performance}, our method achieves the lowest \mae~for acupuncture insertion time and frequency, and matches the best-performing baselines for withdrawal time. Compared to the second-best baselines, our framework reduces the MAE for acupuncture insertion time from 3.26~s to 3.12~s (4.3\% improvement) and for acupuncture frequency from 0.51~Hz to 0.49~Hz (3.9\%). Notably, these improvements are achieved over recent transformer-based AQA frameworks such as PGT, PHI, and Uni-FineParser, which model long-term temporal dynamics but rely on single-view global representations. For withdrawal time, our method matches the best-performing baseline with an MAE of 0.94~s, while exhibiting a comparable standard deviation ($\pm0.14$) to other strong methods. For \REV{Tuina} frequency, PHI achieves the lowest MAE (0.63~Hz), while our method attains 0.64~Hz with reduced standard deviation ($\pm0.05$ vs. $\pm0.07$), indicating improved stability.

Table~\ref{tab:quantitative_performance} further shows that these performance gains are accompanied by stable prediction behavior. In particular, our method exhibits lower standard deviation for \REV{Tuina} frequency ($\pm0.05$) compared to PHI ($\pm0.07$), and comparable standard deviation for withdrawal time, indicating that the observed improvements are not driven by outliers.
The \REV{lower mean errors than most recent methods} suggest that explicitly modeling localized hand-centric interactions and structured multi-view cues \REV{could contribute to favorable metrological performance} in \tcm~rehabilitation.
Although the absolute prediction errors remain non-trivial due to the inherent variability and complexity of fine-grained hand motions in \tcm~procedures, our approach still \REV{achieves the best or comparable mean performance across the evaluated metrological indicators}, demonstrating its potential to provide more precise and reliable metrological assessments in clinical training applications.

\begin{table*}[h]
\scriptsize
\centering
\caption{Performance comparison of Acupuncture skills on the TCM-AQA61-A dataset (Part I: Needle handling and insertion). Results are reported as Accuracy/Weighted F1 (\%, mean$\pm$SD over folds). For each metric, the method with the highest mean is highlighted in bold; when means are identical, the method with the lower SD is selected.}
\label{tab:acupuncture_performance_part1}
\begin{tabular}{lcccc}
\hline
 & \makecell{Needle\\ Holding ($\uparrow$)} 
 & \makecell{Needle\\ Angle ($\uparrow$)} 
 & \makecell{Needle\\ Depth ($\uparrow$)} 
 & \makecell{Quick Needle\\ Insertion ($\uparrow$)} \\
\hline

\hline
STGCN \cite{yan2018spatial}
& 0.90$\pm$0.09/\textbf{0.85}$\pm$0.08
& 0.62$\pm$0.11/0.55$\pm$0.11
& 0.57$\pm$0.13/0.41$\pm$0.15
& 0.57$\pm$0.18/0.51$\pm$0.16 \\

STNN \cite{liao2020deep}
& 0.90$\pm$0.08/0.85$\pm$0.12
& \textbf{0.68}$\pm$0.05/0.56$\pm$0.07
& 0.57$\pm$0.10/0.41$\pm$0.16
& 0.54$\pm$0.17/0.37$\pm$0.17 \\

STGCN-LSTM \cite{deb2022graph}
& 0.90$\pm$0.08/0.85$\pm$0.12
& 0.68$\pm$0.08/0.56$\pm$0.11
& 0.57$\pm$0.07/0.41$\pm$0.07
& 0.42$\pm$0.10/0.42$\pm$0.10 \\

STGCN-RI \cite{zheng2023skeleton}
& 0.86$\pm$0.13/0.83$\pm$0.10
& 0.59$\pm$0.05/0.54$\pm$0.06
& 0.52$\pm$0.08/0.52$\pm$0.09
& 0.49$\pm$0.17/0.47$\pm$0.16 \\

FineParser \cite{xu2024fineparser}
& 0.88$\pm$0.08/0.85$\pm$0.13
& 0.63$\pm$0.09/0.57$\pm$0.14
& 0.52$\pm$0.06/0.46$\pm$0.07
& 0.55$\pm$0.07/0.51$\pm$0.07 \\

Uni-FineParser \cite{xu2025human}
& 0.90$\pm$0.10/0.85$\pm$0.14
& 0.68$\pm$0.14/0.56$\pm$0.18
& 0.59$\pm$0.07/0.48$\pm$0.11
& 0.50$\pm$0.10/0.47$\pm$0.16 \\

PGT \cite{zhang2025pose}
& 0.90$\pm$0.09/0.85$\pm$0.13
& 0.66$\pm$0.10/0.60$\pm$0.15
& 0.47$\pm$0.08/0.39$\pm$0.14
& 0.52$\pm$0.08/0.48$\pm$0.11 \\

PHI \cite{zhou2025phi}
& \textbf{0.90}$\pm$0.06/0.85$\pm$0.09
& 0.68$\pm$0.10/0.56$\pm$0.12
& 0.55$\pm$0.14/0.42$\pm$0.16
& 0.53$\pm$0.15/0.44$\pm$0.20 \\

\hline
Ours
& 0.88$\pm$0.08/0.85$\pm$0.13
& 0.63$\pm$0.12/\textbf{0.61}$\pm$0.14
& \textbf{0.60}$\pm$0.08/\textbf{0.60}$\pm$0.08
& \textbf{0.63}$\pm$0.12/\textbf{0.63}$\pm$0.11 \\
\hline
\end{tabular}
\end{table*}

\begin{table*}[h]
\scriptsize
\centering
\caption{Performance comparison of Acupuncture skills on the TCM-AQA61-A dataset (Part II: Manipulation and withdrawal). Results are reported as Accuracy/Weighted F1 (\%, mean$\pm$SD over folds). For each metric, the method with the highest mean is highlighted in bold; when means are identical, the method with the lower SD is selected.}
\label{tab:acupuncture_performance_part2}
\begin{tabular}{lccccc}
\hline
 & \makecell{Lifting \& Thrusting\\ Frequency ($\uparrow$)}
 & \makecell{Lifting \& Thrusting\\ Amplitude ($\uparrow$)}
 & \makecell{Twisting\\ Frequency ($\uparrow$)}
 & \makecell{Twisting\\ Amplitude ($\uparrow$)}
 & \makecell{Quick Needle\\ Withdrawal ($\uparrow$)} \\
\hline
STGCN \cite{yan2018spatial}
& 0.68$\pm$0.10/\textbf{0.65}$\pm$0.07
& 0.70$\pm$0.19/0.63$\pm$0.22
& 0.83$\pm$0.08/0.76$\pm$0.11
& 0.93$\pm$0.11/0.90$\pm$0.10
& 0.85$\pm$0.08/0.80$\pm$0.11 \\

STNN \cite{liao2020deep}
& 0.75$\pm$0.11/0.64$\pm$0.14
& 0.65$\pm$0.18/0.51$\pm$0.24
& \textbf{0.83}$\pm$0.05/0.76$\pm$0.07
& \textbf{0.93}$\pm$0.06/\textbf{0.90}$\pm$0.09
& 0.86$\pm$0.10/0.80$\pm$0.14 \\

STGCN-LSTM \cite{deb2022graph}
& \textbf{0.75}$\pm$0.10/0.64$\pm$0.11
& 0.65$\pm$0.14/0.51$\pm$0.17
& \textbf{0.83}$\pm$0.05/0.76$\pm$0.07
& \textbf{0.93}$\pm$0.06/\textbf{0.90}$\pm$0.09
& 0.86$\pm$0.10/0.80$\pm$0.14 \\

STGCN-RI \cite{zheng2023skeleton}
& 0.62$\pm$0.13/0.60$\pm$0.12
& 0.49$\pm$0.20/0.47$\pm$0.18
& 0.77$\pm$0.04/0.72$\pm$0.04
& 0.93$\pm$0.10/0.90$\pm$0.11
& \textbf{0.88}$\pm$0.08/\textbf{0.84}$\pm$0.12 \\

FineParser \cite{xu2024fineparser}
& 0.70$\pm$0.13/0.64$\pm$0.16
& 0.59$\pm$0.21/0.58$\pm$0.19
& 0.82$\pm$0.12/0.76$\pm$0.18
& \textbf{0.93}$\pm$0.06/\textbf{0.90}$\pm$0.09
& 0.78$\pm$0.13/0.78$\pm$0.16 \\

Uni-FineParser \cite{xu2025human}
& 0.75$\pm$0.11/0.65$\pm$0.14
& 0.75$\pm$0.11/0.73$\pm$0.13
& 0.83$\pm$0.14/0.77$\pm$0.19
& \textbf{0.93}$\pm$0.06/\textbf{0.90}$\pm$0.09
& 0.85$\pm$0.10/0.80$\pm$0.15 \\

PGT \cite{zhang2025pose}
& \textbf{0.75}$\pm$0.10/0.65$\pm$0.14
& 0.62$\pm$0.09/0.63$\pm$0.08
& 0.83$\pm$0.13/0.77$\pm$0.19
& \textbf{0.93}$\pm$0.06/\textbf{0.90}$\pm$0.09
& 0.80$\pm$0.08/0.78$\pm$0.14 \\

PHI \cite{zhou2025phi}
& 0.75$\pm$0.18/0.65$\pm$0.23
& \textbf{0.77}$\pm$0.09/\textbf{0.74}$\pm$0.10
& 0.82$\pm$0.06/0.75$\pm$0.09
& \textbf{0.93}$\pm$0.06/\textbf{0.90}$\pm$0.09
& 0.86$\pm$0.11/0.81$\pm$0.16 \\

\hline
Ours
& 0.65$\pm$0.09/0.60$\pm$0.08
& 0.75$\pm$0.08/\textbf{0.74}$\pm$0.10
& 0.83$\pm$0.12/\textbf{0.78}$\pm$0.17
& \textbf{0.93}$\pm$0.06/\textbf{0.90}$\pm$0.09
& 0.85$\pm$0.11/0.80$\pm$0.16 \\
\hline
\end{tabular}
\end{table*}

\begin{table*}[ht]
\scriptsize
\centering
\caption{Performance comparison of Tuina skills on the TCM-AQA61-T dataset. Results are reported as Accuracy/Weighted F1 (\%, mean$\pm$SD over folds). For each metric, the method with the highest mean is highlighted in bold; when means are identical, the method with the lower SD is selected.}
\label{tab:tuina_performance}
\begin{tabular}{lcccccc}
\hline
 & \makecell{Standard\\ Action ($\uparrow$)}
 & \makecell{Hollow\\ Palm ($\uparrow$)}
 & \makecell{Solid\\ Fingers ($\uparrow$)}
 & \makecell{Slow\\ Movement ($\uparrow$)}
 & Depth ($\uparrow$)
 & Frequency ($\uparrow$) \\
\hline
STGCN \cite{yan2018spatial}
& 0.68$\pm$0.05/0.66$\pm$0.06
& 0.66$\pm$0.14/0.64$\pm$0.15
& 0.94$\pm$0.06/0.92$\pm$0.07
& \textbf{0.98}$\pm$0.03/\textbf{0.97}$\pm$0.05
& 0.82$\pm$0.05/0.85$\pm$0.05
& 0.47$\pm$0.25/0.47$\pm$0.26 \\

STNN \cite{liao2020deep}
& 0.61$\pm$0.07/0.46$\pm$0.12
& 0.71$\pm$0.05/0.60$\pm$0.05
& 0.94$\pm$0.04/0.92$\pm$0.06
& \textbf{0.98}$\pm$0.03/\textbf{0.97}$\pm$0.05
& 0.94$\pm$0.04/0.92$\pm$0.06
& 0.42$\pm$0.11/0.36$\pm$0.15 \\

STGCN-LSTM \cite{deb2022graph}
& 0.50$\pm$0.14/0.46$\pm$0.20
& 0.71$\pm$0.13/0.60$\pm$0.18
& 0.94$\pm$0.04/0.92$\pm$0.06
& \textbf{0.98}$\pm$0.03/\textbf{0.97}$\pm$0.05
& 0.94$\pm$0.04/\textbf{0.97}$\pm$0.06
& 0.43$\pm$0.10/0.43$\pm$0.14 \\

STGCN-RI \cite{zheng2023skeleton}
& 0.56$\pm$0.18/0.56$\pm$0.21
& 0.68$\pm$0.20/0.60$\pm$0.21
& 0.80$\pm$0.06/0.84$\pm$0.06
& 0.98$\pm$0.04/\textbf{0.97}$\pm$0.05
& 0.84$\pm$0.04/0.86$\pm$0.05
& 0.49$\pm$0.06/0.45$\pm$0.06 \\

FineParser \cite{xu2024fineparser}
& 0.66$\pm$0.09/0.64$\pm$0.10
& 0.66$\pm$0.12/0.64$\pm$0.13
& 0.94$\pm$0.04/0.92$\pm$0.06
& 0.98$\pm$0.04/\textbf{0.97}$\pm$0.05
& 0.94$\pm$0.07/0.92$\pm$0.10
& 0.59$\pm$0.08/0.58$\pm$0.06 \\

Uni-FineParser \cite{xu2025human}
& 0.59$\pm$0.10/0.58$\pm$0.11
& 0.64$\pm$0.12/0.65$\pm$0.09
& 0.91$\pm$0.09/0.90$\pm$0.08
& 0.96$\pm$0.04/0.96$\pm$0.05
& 0.91$\pm$0.08/0.90$\pm$0.09
& \textbf{0.63}$\pm$0.04/\textbf{0.62}$\pm$0.04 \\

PGT \cite{zhang2025pose}
& 0.66$\pm$0.07/0.66$\pm$0.07
& 0.70$\pm$0.18/0.67$\pm$0.21
& 0.94$\pm$0.04/0.92$\pm$0.06
& \textbf{0.98}$\pm$0.03/\textbf{0.97}$\pm$0.05
& 0.93$\pm$0.08/0.91$\pm$0.10
& 0.61$\pm$0.10/0.56$\pm$0.10 \\

PHI \cite{zhou2025phi}
& 0.60$\pm$0.13/0.51$\pm$0.19
& 0.72$\pm$0.10/0.61$\pm$0.13
& \textbf{0.95}$\pm$0.07/0.92$\pm$0.10
& \textbf{0.98}$\pm$0.03/\textbf{0.97}$\pm$0.05
& \textbf{0.95}$\pm$0.04/0.92$\pm$0.06
& 0.62$\pm$0.11/0.55$\pm$0.11 \\

\hline
Ours
& \textbf{0.71}$\pm$0.15/\textbf{0.70}$\pm$0.15
& \textbf{0.73}$\pm$0.14/\textbf{0.69}$\pm$0.18
& 0.94$\pm$0.02/\textbf{0.92}$\pm$0.05
& \textbf{0.98}$\pm$0.03/\textbf{0.97}$\pm$0.05
& 0.94$\pm$0.01/0.92$\pm$0.10
& 0.59$\pm$0.05/0.59$\pm$0.04 \\

\hline
\end{tabular}
\end{table*}

\subsection{Classification-Based Assessment Performance}
\noindent\textbf{Baselines.}
We employed the same baseline methods as in the metrological assessment experiments. However, PHI~\cite{zhou2025phi} and PGT~\cite{zhang2025pose} were originally formulated for continuous score regression. To enable fair comparison in the classification-based evaluation, we replaced their regression heads with categorical prediction layers while keeping the backbone architectures unchanged.  All other network configurations remained consistent with their original implementations.

\noindent\textbf{Results.}
As practical metrological assessments may not fully reflect practice quality, we also conducted classification-based evaluations. The results presented in Tables~\ref{tab:acupuncture_performance_part1},~\ref{tab:acupuncture_performance_part2} and~\ref{tab:tuina_performance} demonstrate that our proposed \cmeaqa~framework achieves competitive or superior performance relative to the strongest competing baselines across both acupuncture and \REV{Tuina} tasks. Importantly, these comparisons include recent transformer-based AQA frameworks such as Uni-FineParser, PGT, and PHI, which employ global spatio-temporal modeling but do not explicitly incorporate structured multi-view alignment or localized hand-centric fusion.

For acupuncture, our method delivers improvements in several key skills. For example, in Needle Depth, our framework achieves an F1 score of 0.60, representing a relative gain of approximately 15\% compared with the second-best method (0.52). This improvement is particularly notable given that transformer-based methods such as PGT and PHI achieve lower F1 scores (0.39 and 0.42, respectively), suggesting that global temporal modeling alone is insufficient for fine-grained depth discrimination. In Quick Needle Insertion, we obtain $0.63/0.63$ in accuracy/F1 score, which is about 6\% higher in accuracy and 12\% higher in F1 score than the second-best baseline ($0.57/0.51$). This margin is maintained over both graph-based and transformer-based competitors, indicating stronger modeling of rapid hand-object interactions. For Twisting Frequency, our model matches the highest accuracy (0.83) while slightly improving the F1 score (0.78 vs.\ 0.76). These gains are accompanied by \REV{moderate fold-level variability}. In Needle Depth, our method achieves $0.60\pm0.08$ F1, exhibiting substantially lower standard deviation than PHI ($0.42\pm0.16$) and comparable stability to the strongest graph-based baselines.
Similarly, for Quick Needle Insertion, our F1 standard deviation ($\pm0.11$) is lower than that of STGCN ($\pm0.16$), \REV{suggesting that the observed mean-level advantage is less likely to be driven by a single outlier split.}

In \REV{Tuina} assessment, our method also outperforms not only classical pose-based baselines but also advanced transformer models on key indicators such as Standard Action and Hollow Palm. For example, compared with PHI (0.51 F1) and Uni-FineParser (0.58 F1) in Standard Action, our method achieves $0.70\pm0.15$ F1, representing a substantial relative improvement. Similarly, in Hollow Palm, our framework achieves $0.69\pm0.18$ F1, exceeding PHI (0.61) and PGT (0.67). 
This \REV{performance is supported by} the fusion of visual and skeletal features and the \REV{integration} of egocentric-view cues, which \REV{provide complementary information for fine-grained motion modeling and occlusion mitigation when visual hand-object cues are available in the TCM setting}. 
These findings suggest that explicitly modeling localized hand-centric interactions and cross-view complementary information \REV{is associated with favorable mean performance over purely global transformer modeling strategies on several fine-grained TCM rehabilitation indicators}. 
These results further highlight the potential of our \cmeaqa~framework to provide convenient yet accurate feedback for \REV{structured} \tcm~rehabilitation training.

We further examine whether the observed improvements could be attributed to label imbalance, as most participants are medical students and several indicators exhibit moderate skew (Fig.~\ref{fig:hist}). 
Notably, the performance gains of CME-AQA are not confined to near-saturated indicators. 
For relatively balanced skills such as Needle Depth (26/35), Quick Needle Insertion (28/33), and Tuina Frequency (31/30), our method \REV{achieves superior or comparable mean performance relative to the strongest baselines} (Tables~\ref{tab:acupuncture_performance_part1}-\ref{tab:tuina_performance}).
Meanwhile, for highly imbalanced indicators such as Twisting Amplitude (4/57) and Slow Movement (60/1), our method achieves performance comparable to or slightly better than competing methods rather than exhibiting artificially inflated accuracy. 
These results suggest that the \REV{observed improvements are not confined to label-skewed indicators and could be associated with improved modeling of fine-grained interactions.}

\begin{table}[h]
\scriptsize
\centering
\caption{Overall paired comparison between CME-AQA (Ours) and PHI across five cross-validation folds on TCM-AQA61-A (Acupuncture) and TCM-AQA61-T (Tuina).
Reported values are fold means over all classes (mean$\pm$SD; higher is better $\uparrow$); $p$-values are from paired fold-level tests ($n{=}5$).}
\label{tab:stat_overall_ours_vs_phi}
\begin{tabular}{l l c c c c}
\hline
\textbf{Dataset} & \textbf{Metric} & $n$ & PHI & Ours & $p$ \\
\hline
\multirow{2}{*}{TCM-AQA61-A}
& Accuracy & 5 & \textbf{0.75}$\pm$0.03 & \textbf{0.75}$\pm$0.03 & 0.73 \\
& Weighted F1       & 5 & 0.68$\pm$0.05 & \textbf{0.72}$\pm$0.04 & 0.35 \\
\hline
\multirow{2}{*}{TCM-AQA61-T}
& Accuracy & 5 & 0.80$\pm$0.04 & \textbf{0.81}$\pm$0.04 & 0.19 \\
& Weighted F1       & 5 & 0.74$\pm$0.05 & \textbf{0.79}$\pm$0.04 & 0.08 \\
\hline
\end{tabular}
\end{table}

\subsection{Overall Fold-Level Statistical Comparison}
Table~\ref{tab:stat_overall_ours_vs_phi} presents an overall paired comparison between CME-AQA (Ours) and PHI~\cite{zhou2025phi}, which \REV{achieves} the strongest or second-best performance among the competing baselines in the classification-based evaluation.
For each cross-validation fold, Accuracy and F1 scores are averaged across all classes to obtain a single fold-level metric, and results are reported as mean $\pm$ standard deviation over folds.
Statistical significance is assessed using a paired two-sided $t$-test at the fold level to evaluate performance differences across splits.
\REV{Given the limited number of folds ($n=5$), the resulting $p$-values are treated as exploratory.}

Overall, CME-AQA \REV{shows higher descriptive mean F1 scores} on both acupuncture and \REV{Tuina} datasets, with \REV{a larger mean difference} observed in \REV{Tuina} assessment.
\REV{The overall fold-level pattern is in line with the per-indicator AQA results reported in Tables~\ref{tab:acupuncture_performance_part1},~\ref{tab:acupuncture_performance_part2}, and~\ref{tab:tuina_performance}. Given that statistical significance is not reached in the paired fold-level tests, these results are best interpreted as descriptive trends rather than statistically confirmed improvements, motivating further validation with larger cohorts.}

\begin{figure}[h]
\centering
\includegraphics[scale=0.13]{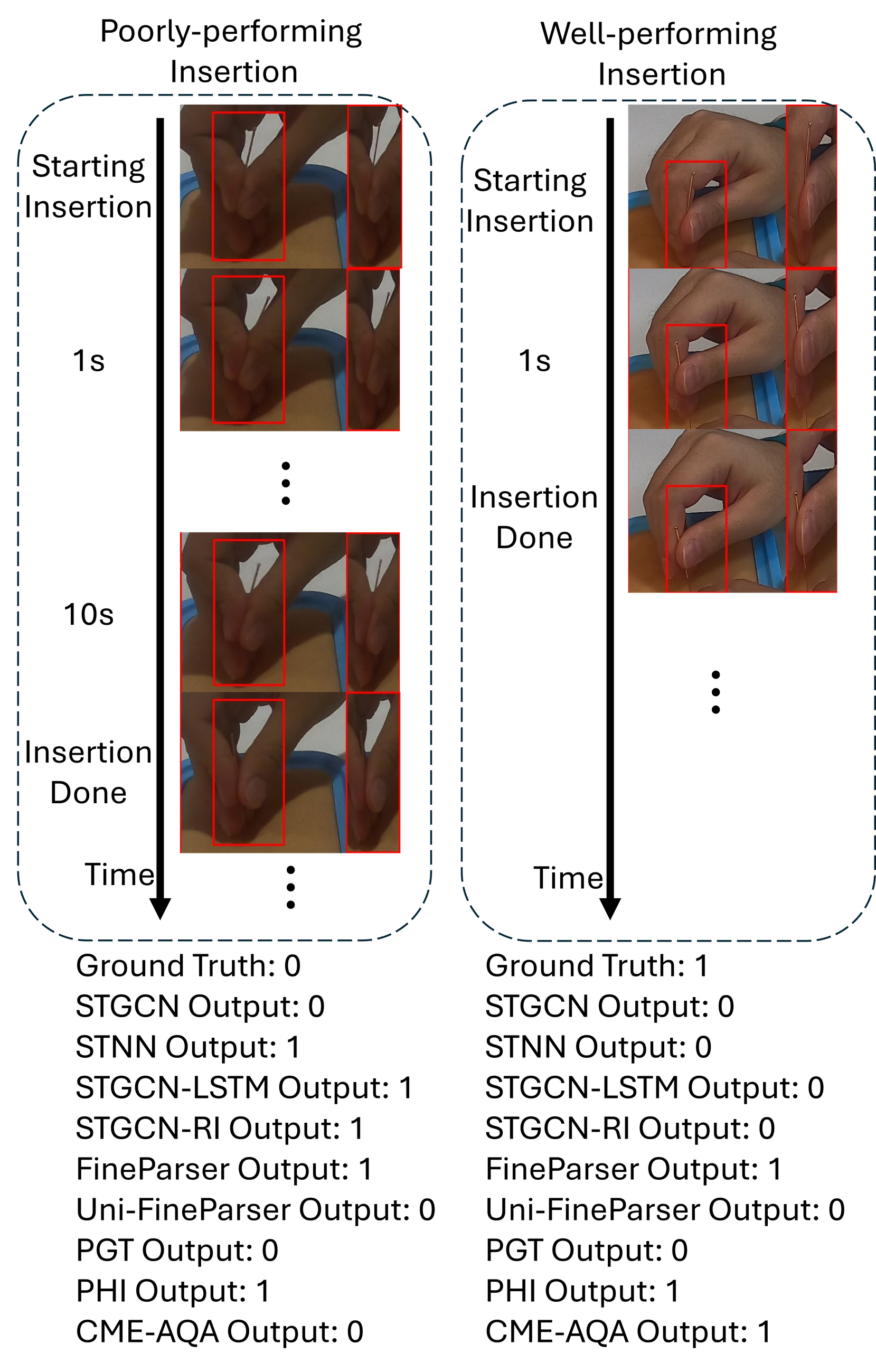}
\caption{Critical case comparison for Quick Needle Insertion. The red box highlights the hand–needle interaction. The well-performing trial completes insertion within 3s, whereas the poorly performing trial lasts about 10s with slower motion. CME-AQA correctly differentiates the two cases, while strong baselines misclassify at least one instance. See supplementary videos for full sequences.}
\label{fig:tcm_case}
\end{figure}

\begin{figure}[h]
\centering
\includegraphics[scale=0.15]{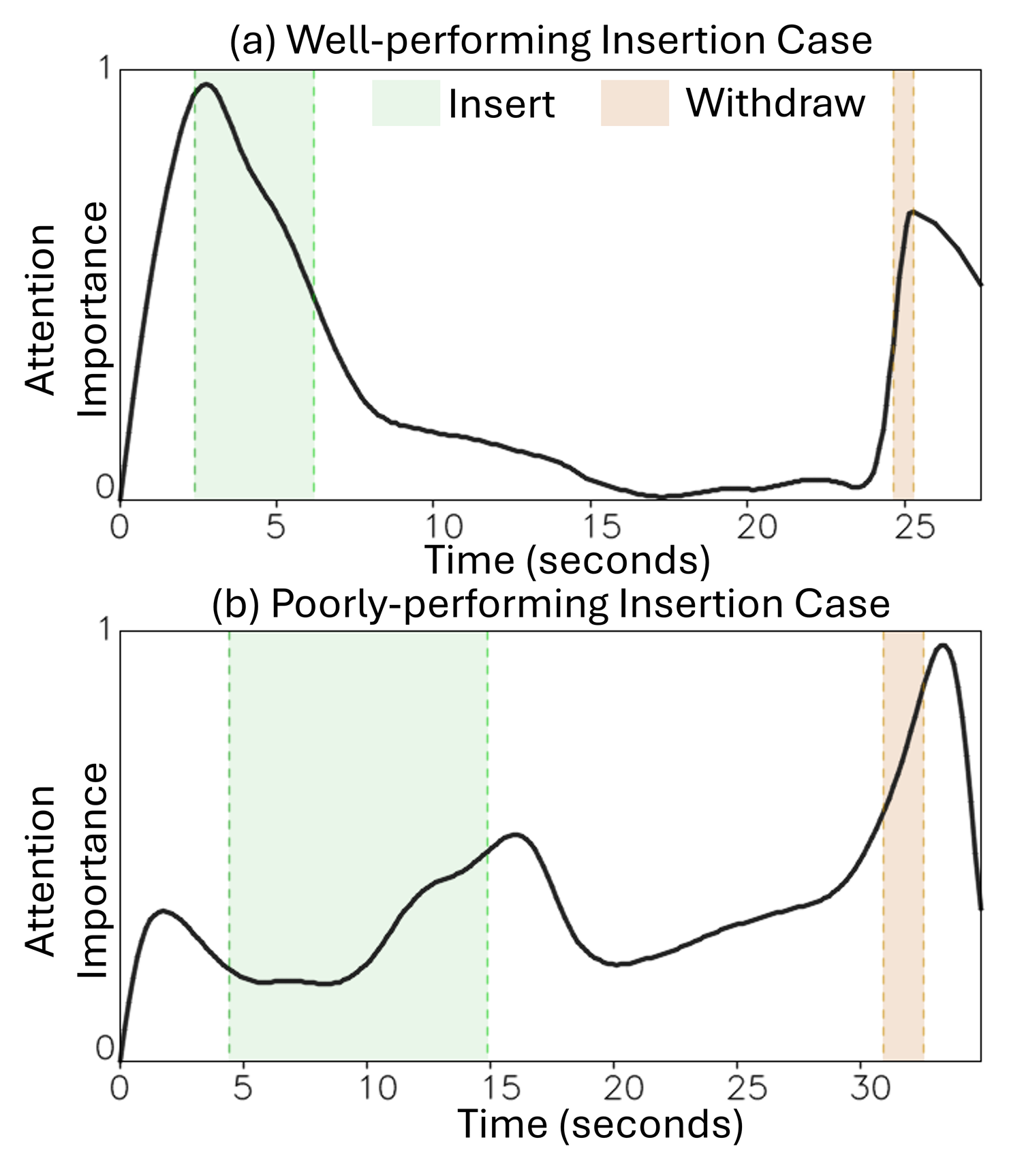}
\caption{Temporal attention patterns learned by CME-AQA for the same well-performing (a) and poorly performing (b) trials. Shaded regions denote insertion (green) and withdrawal (brown) phases. In the well-performing case, attention is sharply concentrated within the brief insertion window, whereas in the poorly performing case it is temporally dispersed across the prolonged insertion period. This behavior indicates that the model captures time-sensitive hand–needle interaction dynamics rather than relying on coarse temporal aggregation.}
\label{fig:tcm_attn}
\end{figure}

\subsection{Qualitative Analysis}
\noindent\textbf{Critical Case Comparison.}
To further examine the effectiveness of CME-AQA under challenging conditions, we present a critical case comparison of Quick Needle Insertion in Fig.~\ref{fig:tcm_case}. This aspect requires rapid and precise hand–needle coordination to achieve therapeutic effects while minimizing discomfort \cite{yin2011high}. The well-performing trial completes insertion within \(\sim\)3s, whereas the poorly performing trial lasts \(\sim\)10s with slower execution. CME-AQA correctly distinguishes the two cases, while strong baselines (FineParser, Uni-FineParser, and PHI) misclassify at least one instance.

\noindent\textbf{Temporal Attention Visualization.}
Figure~\ref{fig:tcm_attn} visualizes the learned temporal attention patterns for the same trials. The cross-attention weights are aggregated across query positions to yield a global temporal importance curve. In the well-performing case, attention exhibits a sharp peak tightly aligned with the annotated insertion window, indicating focused modeling of the brief hand–needle interaction. In contrast, the poorly performing case shows a broader, double-peaked pattern spanning the prolonged insertion period, reflecting extended and less decisive motion. This structured but temporally diffused response suggests that CME-AQA captures key events while remaining sensitive to variations in execution dynamics rather than relying on coarse temporal cues.

\begin{table*}[h]
\scriptsize
\centering
\caption{Ablation study on the TCM-AQA61-A dataset (measured by Accuracy/F1 score). Higher values indicate better performance.}
\begin{tabular}{lcccccccccc}
\hline
 &&\makecell{Needle\\ Holding} & \makecell{Needle\\ Angle} & \makecell{Needle\\ Depth} & \makecell{Quick Needle \\ Insertion} & \makecell{Lifting \& Thrusting \\ Frequency} & \makecell{Lifting \& Thrusting \\Amplitude} & \makecell{Twisting \\ Frequency} & \makecell{Twisting\\ Amplitude} & \makecell{Quick Needle \\ Withdrawal} \\

\hline
\parbox[t]{2mm}{\multirow{5}{*}{\rotatebox[origin=c]{90}{\makecell{Modality \\\& View}}}}&Ours &\textbf{0.88/0.85} & \textbf{0.63/0.61} &\textbf{0.60/0.60}&0.63/0.63 &0.65/0.60 & \textbf{0.75/0.74}&0.83/0.78&\textbf{0.93/0.90} &\textbf{0.85/0.80}\\
&(w/o TPV)&0.85/0.82&0.62/0.52  &0.59/0.58&0.59/0.58 &0.68/0.61  &0.68/0.68 &0.80/0.76&0.91/0.89 &\textbf{0.85}/0.79  \\
&(w/o FPV)& 0.85/0.82&0.55/0.55  &0.55/0.54&\textbf{0.68}/0.63 &0.73/0.60  &0.73/0.73 &\textbf{0.85/0.81}&\textbf{0.93/0.90} &0.83/0.79  \\
&(w/o Visual) & 0.86/0.83&0.59/0.58  &\textbf{0.60/0.60}&0.65/\textbf{0.65} &0.67/0.62  &0.73/0.72 &\textbf{0.85}/0.79&\textbf{0.93/0.90} &0.83/0.79  \\
&(w/o Pose) & 0.85/0.82&0.59/0.54  &0.55/0.54&0.55/0.55 &\textbf{0.75/0.64}  &0.63/0.62 &0.83/0.76&\textbf{0.93/0.90} &0.83/0.79  \\
\hline
\parbox[t]{2mm}{\multirow{3}{*}{\rotatebox[origin=c]{90}{\makecell{Design}}}}&Ours &\textbf{0.88/0.85} & 0.63/0.61 &\textbf{0.60/0.60}&\textbf{0.63/0.63} &0.65/0.60 & \textbf{0.75/0.74}&\textbf{0.83/0.78}&\textbf{0.93/0.90} &\textbf{0.85/0.80}\\
&(w/o \avpf) & \textbf{0.88/0.85}&\textbf{0.65}/0.61 &0.54/0.53&0.49/0.48 &\textbf{0.72/0.63} &0.72/0.72 &0.81/0.75 &\textbf{0.93/0.90}&0.80/0.77 \\
&\makecell{(w/o \mva)} & \textbf{0.88/0.85}&\textbf{0.65/0.63} &0.44/0.43&0.60/0.60 &0.65/\textbf{0.63} &0.73/0.73 &\textbf{0.83/0.78} &\textbf{0.93/0.90}&0.83/0.79 \\
\hline
\parbox[t]{2mm}{\multirow{4}{*}{\rotatebox[origin=c]{90}{\makecell{Share\\Weight}}}}&Ours  &\textbf{0.88/0.85} & \textbf{0.63/0.61} &\textbf{0.60/0.60}&\textbf{0.63/0.63} &0.65/\textbf{0.60} &0.75/0.74 &\textbf{0.83/0.78}&\textbf{0.93/0.90} &\textbf{0.85/0.80}\\
&Late&\textbf{0.88/0.85} & 0.62/0.60 &0.59/0.58&0.57/0.56 &0.63/\textbf{0.60} &\textbf{0.78/0.77} &\textbf{0.83/0.78}&0.93/0.90 &0.83/0.79\\
&No share  & 0.86/0.83&\textbf{0.63}/0.60  &0.57/0.54&0.52/0.51 &0.63/\textbf{0.60}  &0.75/0.75 &0.81/0.77&0.93/0.90 &0.83/0.79  \\
&All share  & 0.86/0.83&0.62/0.56  &0.59/0.58&0.54/0.53 &\textbf{0.67/0.60}  &0.77/0.74 &\textbf{0.83}/0.76&0.93/0.90 &0.83/0.79  \\
\hline
\label{tab:ablation_acupuncture}
\end{tabular}
\end{table*}

\subsection{Ablation Study}
To validate the effectiveness of the proposed components, we conducted a comprehensive ablation study across (i) view/modality settings, (ii) module design, and (iii) weight-sharing strategies; see Table~\ref{tab:ablation_acupuncture}.

\noindent\textbf{View and Modality Ablation.}
We removed each view (egocentric/FPV or exocentric/TPV) and each modality (visual or pose) to assess their contributions. Integrating both views and both modalities yields the strongest results. Removing FPV leads to noticeable drops in accuracy and F1 for fine hand-movement aspects (e.g., needle depth), underscoring the role of multiple viewpoints in mitigating self-occlusion. Omitting pose features degrades tasks that require precise spatial reasoning (e.g., Needle Depth). These findings indicate that visual and pose cues are complementary.

\noindent\textbf{Framework Design Ablation.}
Replacing the cross-attention in \avpf~with fully connected layers reduces performance, especially for Needle Angle. This indicates that cross-modal attention is necessary to fuse pose and visual information effectively. Removing \mva~and aligning only at the final output also degrades performance, with the largest drop on temporally structured aspects such as Twisting Frequency. These results support the benefit of multi-scale view alignment within the feature hierarchy.

\noindent\textbf{Weight-Sharing Strategies.}
We compared three strategies: (1) share weights in \emph{self-attention only}, (2) \emph{no} weight sharing, and (3) share weights in \emph{both} cross- and self-attention. Our default design—sharing weights in cross-attention while using view-specific self-attention—achieves the best overall accuracy and F1. This suggests cross-attention benefits from shared parameters to capture view-invariant correlations, whereas self-attention benefits from view-specific adaptation to refine task-relevant temporal features.

\begin{figure}[t]
    \centering
    \includegraphics[width=1\linewidth]{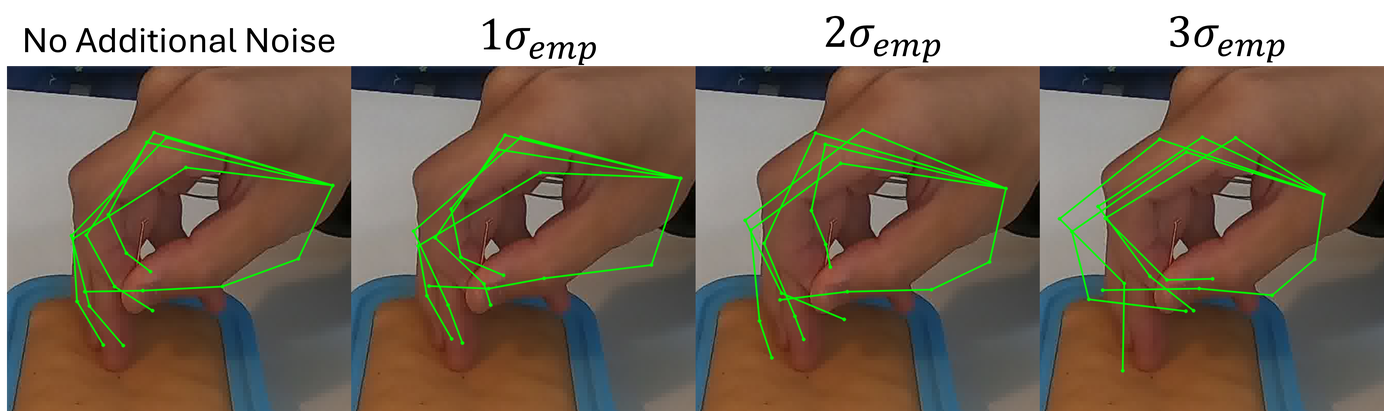}
    \caption{Illustration of progressive pose perturbation. From left to right: original pose (no additional noise), and injected Gaussian jitter with $\sigma = 1\sigma_{\text{emp}}, 2\sigma_{\text{emp}}, 3\sigma_{\text{emp}}$. Noise is added independently to each joint coordinate in normalized space. As $\sigma$ increases, joint locations exhibit progressively larger spatial deviations, leading to increasing geometric distortion of finger alignment and hand contour while preserving overall pose structure at lower perturbation levels.}
    \label{fig:pose_noise}
\end{figure}

\begin{table}[t]
\centering
\scriptsize
\caption{Sensitivity analysis of overall performance on TCM-AQA61-A (Acupuncture). Reported values are fold means over all assessment indicators (mean$\pm$SD; higher is better $\uparrow$; $n{=}5$). Pose noise denotes additive Gaussian jitter on joint coordinates, and multi-view scale refers to the number of aligned feature scales in MVA.}
\label{tab:sensitivity_overall}
\begin{tabular}{lcc}
\hline
Setting &Accuracy & Weighted F1 \\
\hline
\multicolumn{3}{l}{\textit{Pose noise (Gaussian jitter), $\epsilon \sim \mathcal{N}(0,\sigma^2)$}} \\
No additional noise  &  \textbf{0.75}$\pm$0.03 & \textbf{0.72}$\pm$0.04 \\
$1\sigma_{\text{emp}}$  & 0.74$\pm$0.05 & 0.70$\pm$0.05 \\
$2\sigma_{\text{emp}}$& 0.74$\pm$0.05 & 0.69$\pm$0.05 \\
$3\sigma_{\text{emp}}$ & 0.72$\pm$0.05 & 0.67$\pm$0.06 \\
\hline
\multicolumn{3}{l}{\textit{Multi-view alignment scale (number of aligned feature scales in MVA)}} \\
1-scale (early only)        & 0.74$\pm$0.03 & 0.65$\pm$0.03 \\
1-scale (mid only)          & 0.74$\pm$0.03 & 0.64$\pm$0.04\\
1-scale (late only)         & 0.73$\pm$0.03 & 0.70$\pm$0.04 \\
2-scale (early+mid)         & 0.74$\pm$0.03 & 0.65$\pm$0.04 \\
2-scale (early+late)        & 0.74$\pm$0.03 & 0.67$\pm$0.05 \\
2-scale (mid+late)          & \textbf{0.75}$\pm$0.03 & 0.67$\pm$0.05 \\
3-scale (all scales, default MVA) & \textbf{0.75}$\pm$0.03 & \textbf{0.72}$\pm$0.04 \\
\hline
\end{tabular}
\end{table}

\subsection{Sensitivity Analysis}
To better understand the mechanisms of \avpf~and \mva, we conduct a sensitivity analysis examining two factors: (i) pose quality and (ii) the scale of multi-view feature alignment.
\noindent\textbf{Pose noise.}
Using the fully trained model, we inject zero-mean Gaussian noise into pose coordinates to simulate landmark jitter, a common data augmentation strategy~\cite{liu2025systematic}, defining $p' = p + \epsilon$ with $\epsilon \sim \mathcal{N}(0,\sigma^2)$. The empirical median frame-to-frame pose displacement across all sequences is $\sigma_{\text{emp}} = 0.01$ in normalized coordinates. Relative to the median hand size in our dataset, this corresponds to $\sim$7\% of hand width (4\% of hand height). As illustrated in Fig.~\ref{fig:pose_noise}, we evaluate $\sigma \in \{1,2,3\}\sigma_{\text{emp}}$ to simulate different noise levels.

As shown in Table~\ref{tab:sensitivity_overall}, performance remains relatively stable at $1\sigma_{\text{emp}}$, while larger perturbations lead to gradual declines in both metrics. 
This \REV{relative stability under the tested perturbations could be related to the design of \avpf}, where pose features serve as fixed geometric references in cross-attention and are not iteratively updated, thereby reducing the risk of noise amplification across layers.
This behavior is further illustrated in Fig.~\ref{fig:pose_noise}, where stronger perturbations progressively distort joint configurations, weakening the spatial priors used for visual refinement. 
The smooth degradation suggests that the \cmeaqa~framework \REV{remains relatively stable within the simulated lower-to-moderate pose perturbation range tested here, while larger or systematic pose estimation errors may still propagate to the final AQA output.}

\noindent\textbf{Multi-view alignment scale.}
We evaluate different alignment configurations within MVA by retraining each variant while keeping all other components fixed. The three scales correspond to early cross-attention features, \avpf~outputs, and self-attention output features (Sec.~\ref{sec:mva}). As shown in Table~\ref{tab:sensitivity_overall}, performance generally improves as more scales are aligned, and the full configuration achieves the best results. Notably, late-only alignment, performed on self-attention features, already provides competitive performance. This trend reflects the rationale of \mva, indicating that high-level semantic consistency in late features is the primary driver of multi-view robustness. Early and mid-level alignment operate on the shared multimodal backbone features and mainly provide geometric regularization. Aligning all scales combines geometric stabilization with semantic alignment, resulting in more robust view-invariant embeddings.

\label{sec:cpr}
\begin{table}[h]
\scriptsize
\centering
\caption{Mean Absolute Error (Human Experts vs. Our \cmeaqa Framework). Lower values indicate better performance.}
\begin{tabular}{lcccc}
\hline
\textbf{Aspect} & \textbf{Expert 1} & \textbf{Expert 2} &\textbf{MV-STGCN \cite{constable2024advancing}} & \textbf{Ours} \\
\hline
Hand Position & 1.62 & 1.08 &\textbf{0.33}& \textbf{0.33} \\
Arm Position & 0.70 & 0.15 & \textbf{0.07}& \textbf{0.07} \\
Shoulder Position & 0.40 & 0.34 & \textbf{0.13}& \textbf{0.13} \\
Depth & 0.49 & \textbf{0.30}&  0.69& 0.69  \\
Rate & 0.89 & \textbf{0.11} & 1.67&1.78 \\
Compression Release & 1.04 & \textbf{0.98}  &1.00& 1.00 \\
\hline
\end{tabular}
\label{tab:cpr_aqa}
\end{table}
\subsection{Generalization Analysis for \cpr~Skill Assessment}
The generalization results are reported in Table~\ref{tab:cpr_aqa}. Expert~1 and Expert~2 denote the mean absolute error (MAE) of each expert relative to the consensus scores. Compared with the multi-view baseline MV-STGCN~\cite{constable2024advancing}, our single-view framework achieves comparable errors across several aspects: it matches the baseline on Hand Position (0.33), Arm Position (0.07), Shoulder Position (0.13), and Compression Release (1.00); it is identical on Depth (0.69) and slightly higher on Rate (1.78 vs.\ 1.67). Relative to the human experts, our errors are substantially lower on Hand Position, Arm Position, and Shoulder Position. Our error is comparable to that of Expert~2 on Compression Release (1.00 vs.\ 0.98). Errors are higher than those of Expert~2 on Depth and Rate. Notably, although our framework uses only single-view input at inference, it achieves performance comparable to MV-STGCN, which relies on multi-view data at inference.

The higher errors in Rate relative to MV-STGCN~\cite{constable2024advancing} and the higher errors in Depth and Compression Release compared with human experts therefore expose boundary conditions of the proposed method under restricted input modalities. 
In the CPR dataset, inference was conducted using pose information only, as privacy blurring removed reliable RGB cues, thereby limiting the visual input for \avpf~and the interaction cues required for cross-view alignment in \mva. \REV{Therefore, the reduced interaction modeling in CPR reflects a missing-modality boundary condition, rather than negating the intended role of AVPF in TCM procedures where visual hand-object cues are available.}
This restriction prevents the framework from fully exploiting its multi-view interaction modeling capability under single-view input, which partly explains why our Rate is lower than that of MV-STGCN, which directly leverages multi-view observations as input. Moreover, the key distinction between CPR and the TCM task lies not in motion frequency, but in the type of interaction. CPR compression depth and recoil depend on the deformation of an external compliant object (i.e., the manikin chest)~\cite{krasteva2011audiovisual}, rather than on hand pose alone. Such deformable contact dynamics are not directly observable from skeletal sequences, which partly explains why our errors are higher than those of human experts when the relevant interaction dynamics cannot be inferred from pose alone. 
These findings indicate that the \cmeaqa~framework \REV{shows preliminary transferability to pose-dominant criteria}, but its interaction understanding and view alignment capabilities are inherently constrained when critical interaction cues are absent.

\section{Discussion and Conclusion}
In this paper, we presented the \CMEAQA~framework, a vision-based approach that integrates pose–visual feature fusion and cross-view learning for \tcm~rehabilitation AQA. 
To our knowledge, this is the first framework to combine these elements for fine-grained assessment in traditional medicine training, providing a practical tool to support skill development for \REV{learners in structured TCM rehabilitation training}.
We also introduced two multi-view datasets, TCM-AQA61-A (Acupuncture) and TCM-AQA61-T (Tuina), which include recordings from 61 subjects performing representative \tcm~rehabilitation procedures. Quantitative experiments on our datasets demonstrated that our approach achieved over 10\% higher F1 scores in key tasks such as Needle Depth and Quick Needle Insertion. 
Evaluation on the \cpr~dataset~\cite{constable2024advancing} further showed performance comparable to human experts \REV{on several posture-based criteria, suggesting applicability to related structured simulated clinical skill assessments where participant motion is central to evaluation.}

Several limitations of CME-AQA remain. One \REV{important consideration} is the dependency of our framework on the accuracy of initial pose estimation from raw video data. 
\REV{Although the sensitivity analysis in Table~\ref{tab:sensitivity_overall} shows relative stability within the simulated lower-to-moderate perturbation range, this finding does not remove the framework's dependency on pose-estimation quality; larger or systematic pose errors could still propagate to the final AQA output.}
Future work could integrate more robust pose estimation techniques, such as temporal-based pose estimation \cite{wen2023hierarchical}. This integration would help pose estimation align more closely with the demands of our AQA tasks, particularly in ensuring consistent pose estimation for complex hand movements in TCM therapy scenarios.

Another challenge is how objects are semantically represented in the model. Although our \cmeaqa~framework introduces visual features to provide a more comprehensive understanding of the environment, this may distract the focus from motion itself. \REV{The AVPF module is designed to strengthen interaction-aware fusion when visual hand-object cues are available, while settings with missing or unreliable RGB input may require additional sensing or explicit object representations.}
Recent advancements in interaction detection, such as the use of bounding box detection \cite{qiao2022geometric}, segmentation methods \cite{zhu2024geometric}, and dynamic graph representations \cite{zhang2024adaptive} to highlight objects during video analysis, could help address this issue. These methods could emphasize object interactions and provide a more robust representation in our future work.

Finally, although the current dataset provides a valuable benchmark for fine-grained AQA, it reflects a structured teaching cohort in which most participants are medical students. Consequently, certain advanced or rare skill indicators exhibit label imbalance, while intermediate proficiency dominates overall score distributions. 
Although our analysis shows that \REV{the favorable mean-level trends are not confined to near-saturated indicators}, \REV{the present validation should be interpreted within structured TCM rehabilitation training rather than as evidence of generalization across the full spectrum of expertise levels.} \REV{In addition, the overall fold-level comparisons should be regarded as descriptive rather than statistically confirmed improvements.} Future work will expand data collection to include more experienced practitioners and a wider range of error patterns, enabling \REV{more comprehensive evaluation across diverse skill levels and stronger statistical validation with larger cohorts.}

\section*{Acknowledgment}
\REV{This research was supported in part by Capital's Funds for Health Improvement and Research (Ref. 2024-4-21211), the Training Plan for High Level Public Health Technical Talents Construction Project (Ref. TTL-02-40), the Beijing Natural Science Foundation (Ref. L2510128), the National Natural Science Foundation of China (Ref. 8240152532)}, and the EPSRC NortHFutures project (Ref. EP/X031012/1).

\bibliographystyle{IEEEtran} 
\bibliography{cas-refs}

\end{document}